\documentclass{article}

\usepackage[numbers,sort&compress]{natbib}

\usepackage[preprint]{neurips_2021}

\usepackage[utf8]{inputenc} %
\usepackage[T1]{fontenc}    %
\usepackage[colorlinks]{hyperref}       %
\usepackage{url}            %
\usepackage{booktabs}       %
\usepackage{amsfonts}       %
\usepackage{nicefrac}       %
\usepackage{microtype}      %
\usepackage{xcolor}         %

\usepackage{graphicx}
\usepackage{tikz}
\usepackage{comment}
\usepackage{amsmath,amssymb} %
\usepackage{accents}
\usepackage{xspace}
\usepackage{tabularx}
\usepackage{multirow}
\usepackage{multicol}
\usepackage{makecell}
\usepackage{graphicx}
\usepackage{arydshln}
\usepackage{wrapfig}
\usepackage[font=small,labelfont=bf]{caption}

\usepackage[accsupp]{axessibility}  %

\usepackage[capitalize]{cleveref}
\crefname{section}{Sec.}{Secs.}
\Crefname{section}{Section}{Sections}
\Crefname{table}{Table}{Tables}
\crefname{table}{Tab.}{Tabs.}

\newcommand{\btheta}{\boldsymbol{\Theta}}
\newcommand{\bphi}{\boldsymbol{\Phi}}
\newcommand{\bsphi}{\boldsymbol{\widetilde{\phi}}}
\newcommand{\bw}{\boldsymbol{W}}

\newcommand{\bb}{\boldsymbol{b}}
\newcommand{\bwt}{\widetilde{\bw}}

\newcommand{\bbt}{\widetilde{\bb}}
\newcommand{\cin}{C_\text{in}}
\newcommand{\cout}{C_\text{out}}
\newcommand{\bpsi}{\boldsymbol{\Psi}}
\newcommand{\lambdai}{\lambda_I}
\newcommand{\lambdau}{\lambda_U}
\newcommand{\lambdas}{\lambda_S}
\newcommand{\ours}{LilNetX}
\newcommand{\bpsiscale}{\boldsymbol{\Psi}_\text{scale}}
\newcommand{\bpsishift}{\boldsymbol{\Psi}_\text{shift}}
\newcommand{\PreserveBackslash}[1]{\let\temp=\\#1\let\\=\temp}
\newcolumntype{C}[1]{>{\PreserveBackslash\centering}p{#1}}
\newcolumntype{R}[1]{>{\PreserveBackslash\raggedleft}p{#1}}
\newcolumntype{L}[1]{>{\PreserveBackslash\raggedright}p{#1}}

\usepackage{float}
\floatstyle{plaintop}
\restylefloat{table}

\usepackage{tikz}
\usetikzlibrary{arrows.meta, backgrounds, calc, chains, fit, matrix, positioning, shadows, shapes, circuits.ee}

\usepackage{pgfplots}
\pgfplotsset{compat=1.14}
\pgfplotsset{every axis/.append style={enlargelimits={abs=0pt},grid,axis lines=left}}
\pgfplotsset{every axis plot/.append style={thick,mark size=1.5pt,line join=bevel,mark options={solid}}}
\pgfplotsset{label style={font=\small}}
\pgfplotsset{tick label style={font=\footnotesize}}
\pgfplotsset{grid style={color=black!10}}
\pgfplotsset{legend style={draw=none,opacity=.85,font=\footnotesize,cells={anchor=west,opacity=1}}}
\pgfplotsset{every non boxed x axis/.style={xtick align=center,shorten <=-.5\pgflinewidth}}
\pgfplotsset{every non boxed y axis/.style={ytick align=center,shorten <=-.5\pgflinewidth}}
\pgfplotsset{every non boxed z axis/.style={ztick align=center,shorten <=-.5\pgflinewidth}}
\pgfplotsset{/pgf/number format/1000 sep={\,}}
\pgfplotsset{ignore zero/.style={%
  #1ticklabel={\ifdim\tick pt=0pt \else\pgfmathprintnumber{\tick}\fi}
}}

\definecolor{dgray}{HTML}{999999}

\makeatletter
\DeclareRobustCommand\onedot{\futurelet\@let@token\@onedot}
\def\@onedot{\ifx\@let@token.\else.\null\fi\xspace}

 \def\vs{\emph{vs}\onedot}
 
\def\iid{i.i.d\onedot} 
\def\etal{\emph{et al}\onedot}
\makeatother

\title{LilNetX: Lightweight Networks with EXtreme Model Compression and Structured Sparsification}

\author{Sharath Girish\\
  University of Maryland, College Park\And
Kamal Gupta\\
  University of Maryland, College Park \And
\qquad\qquad Saurabh Singh\\
  \qquad\qquad Google Research \And
\qquad\qquad   Abhinav Shrivastava\\
  \qquad\qquad   University of Maryland, College Park
}

\begin{document}

\maketitle

\begin{abstract}
  We introduce LilNetX, an end-to-end trainable technique for neural networks that enables learning models with specified accuracy-rate-computation trade-off. Prior works approach these problems one at a time and often require post-processing or multistage training which become less practical and do not scale very well for large datasets or architectures. Our method constructs a joint training objective that penalizes the self information of network parameters in a reparameterized latent space to encourage small model size while also introducing priors to increase structured sparsity in the parameter space to reduce computation. We achieve up to 50\% smaller model size and 98\% model sparsity on ResNet-20 while retaining the same accuracy on the CIFAR-10 dataset as well as 35\% smaller model size and 42\% structured sparsity on ResNet-50 trained on ImageNet, when compared to existing state-of-the-art model compression methods. Code is available at \href{https://github.com/Sharath-girish/LilNetX}{https://github.com/Sharath-girish/LilNetX}.
\end{abstract}

\section{Introduction}
\label{sec:intro}
Recent research in deep neural networks (DNNs) has shown that large performance gains can be achieved on a variety of computer vision tasks simply by employing larger parameter-heavy and computationally intensive architectures~\cite{he2016deep,dosovitskiy2020image}. However, as the DNNs proliferate in the industry, they often need to be trained repeatedly, transmitted over the network to different devices, and need to perform under hardware constraints with minimal loss in accuracy, all at the same time. Hence, finding ways to reduce the storage size of the models on the devices while simultaneously improving their run-time is of utmost importance. This paper proposes a general purpose neural network training framework to jointly optimize the model parameters for accuracy, model size on the disk and computation, on any given task.

Over the last few years, the research on training smaller and efficient DNNs has followed two seemingly parallel tracks with different goals:
One line of work focuses on model compression to deal with the storage and communication network bottlenecks when deploying a large number of models over the air. While they achieve high levels of compression in terms of memory, their focus is not on reducing computation. They either require additional algorithms with some form of post hoc training~\cite{yeom2021pruning} or quantize the network parameters at the cost of network performance~\cite{courbariaux2015binaryconnect,li2016ternary}. The other line of work focuses on reducing computation through various model pruning techniques~\cite{frankle2018lottery}. The focus of these works is to decrease the number of Floating Point Operations (FLOPs) of the network at the inference time, albeit they are also able to achieve some compression due to fewer parameters. Typically, the cost of storing these pruned networks on disk is much higher than the dedicated model compression works.

\begin{figure}[t]

\begin{center}
\resizebox{0.75\linewidth}{!}{

 \begin{tikzpicture}
	\begin{semilogxaxis}[
		height=5.88cm,
		width=8.88cm,
		grid=major,
		log basis x=10,
        xmin=0.8,
        xmax=100,
        ymin=-0.05,
        ymax=0.65,
        xlabel={Reduction in Model Size},
        ylabel={\% Reduction in Slice FLOPs},
        axis line style={-{Stealth[scale=2]}},
        legend style={at={(0.66,0.9)},anchor=north west},
        log ticks with fixed point,
        x tick label style={/pgf/number format/1000 sep=\,},
	]
		
	\draw [dashed, dgray] (0.001,0.516) -- (34,0.516);
	\draw [dashed, dgray] (34,-0.05) -- (34,0.516);
    \node (p1) at (1.8, 0.05) {\small ResNet-50};
	\addplot[mark=x, mark size=4pt,line width=2pt] coordinates {
        (1., 0.)
    };
    
    \node (p2) at (34, 0.566) {\small LilNetX (Ours)};
    \addplot[mark=*, mark size=3pt]  coordinates {
        (34, 0.516)
    };
    
    \node (p3) at (15, 0.05) {\small Oktay \cite{oktay2019scalable}};
    \addplot[mark=square*, mark size=3pt]  coordinates {
        (19, 0.0)
    };

    \node (p3) at (14, 0.304) {\small Wiedemann\cite{wiedemann2019deepcabac}};
    \addplot[mark=square*, mark size=3pt]  coordinates {
        (17, 0.254)
    };

    \node (p3) at (5, 0.541) {\small Frankle \cite{frankle2018lottery}};
    \addplot[mark=triangle*, mark size=4pt]  coordinates {
        (5, 0.491)
    };
    
    \node (p3) at (2.8, 0.342) {\small Lin \cite{lin2019toward}};
    \addplot[mark=triangle*, mark size=4pt]  coordinates {
        (3, 0.392)
    };

	\end{semilogxaxis}%
  \end{tikzpicture}%
}
\end{center}
\caption{Our method jointly optimizes for size on disk (x-axis), as well as slice FLOPs (y-axis). We compare various approaches using the base ResNet-50~\cite{he2016deep} architecture on ImageNet dataset~\cite{deng2009imagenet} and plot FLOPs vs. size for models with similar accuracy. Prior methods optimize for either quantization ($\blacksquare$) or pruning ($\blacktriangle$) objectives. Our approach, \ours, enables training a deep model while simultaneously optimizing for computation (using structured sparsity) as well as model size. Refer \cref{tab:baseline_comparison} for details.}
\label{fig:teaser}
\vspace{-1em}
\end{figure}
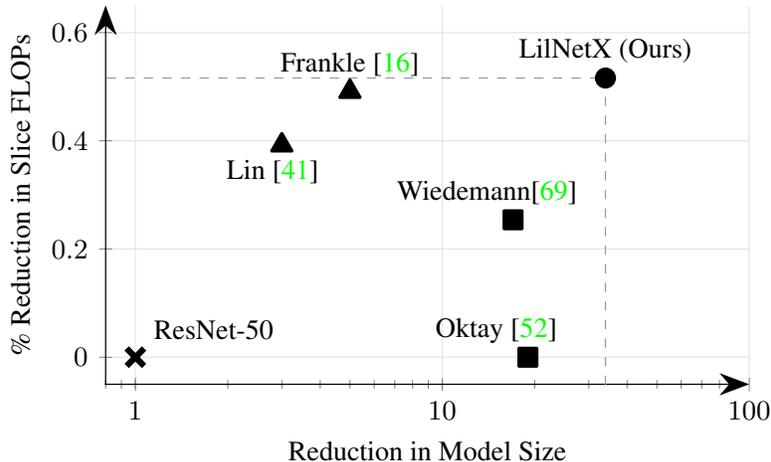

In this work, we bridge the gap between the two and show for the first time that it is indeed possible to jointly optimize the model in terms of both the compression  to reduce disk space as well as structured sparsity to reduce computation (\cref{fig:teaser}). We propose to perform model compression by penalizing the entropy of weights that are quantized in a reparameterized latent space. This idea of reparameterized quantization~\cite{oktay2019scalable} is extremely effective in reducing the effective model size on the disk, however, requires the full dense model during the inference. To address this shortcoming, we introduce key design changes to encourage structured and unstructured parameter sparsity in the model and enable tradeoff with model compression rates and accuracy. Our priors reside in reparameterized latent space while encouraging sparsity in the model space. More specifically, we introduce the notion of slice sparsity, a form of structured sparsity where each $K\times K$ slice is fully zero for a convolutional kernel of filter size $K$. Unlike unstructured sparsity which have irregular memory access and offer little practical speedups, slice-structured sparsity allows for convolutions to be represented as block sparse matrix multiplications which can be exploited for computational gains and inference speedups through dedicated libraries like NVIDIA's CuSparse library \cite{narang2017block}. Slice structured sparsity hits the middle ground between the advantages of high sparsity ratios from unstructured sparsity and practical inference speedups from fully structured sparsity.

Extensive experimentation on three datasets show that our framework achieves extremely high levels of sparsity with little to no performance drops. In addition to increased sparsity, the introduced priors show gains even in model compression compared to Oktay \etal~\cite{oktay2019scalable}. By varying the weight of the priors, we establish a trade-off between model size and accuracy as well as sparsity and accuracy. As our method achieves high levels of model compression along with structured sparsity, we dub it LilNetX - Lightweight Networks with EXtreme Compression and Structured Sparsification. We summarize our contributions below.

\begin{itemize}
    \item We introduce LilNetX, an algorithm to jointly perform model compression as well as structured and unstructured sparsification for direct computational gains in network inference. Our algorithm can be trained end-to-end using a single joint optimization objective and does not require post-hoc training or post-processing.
    \item With extensive ablation studies and results, we show the effectiveness of our approach while outperforming existing approaches in both model compression and pruning in most networks and dataset setups.
\end{itemize}

\section{Related Work}
\label{sec:related}
Typical model compression methods usually follow some form of quantization, parameter pruning, or both. Both lines of work focus on reducing the size of the model on the disk, and/or increasing the speed of the network during the inference time, while maintaining an acceptable level of classification accuracy. In this section, we discuss prominent quantization and pruning techniques.

\subsection{Model Pruning}
It was well established early on that it is possible to prune a large number of neural network weights without significant loss in the performance~\cite{lecun1990optimal,reed1993pruning,han2015learning}. Based on the stage of training at which weights are pruned, we can categorize model pruning methods broadly into three classes.

\textbf{Pruning at initialization} techniques such as SNIP~\cite{lee2018snip}, GraSP~\cite{wang2020picking}, NTT \cite{liu2020finding}, and  SynFlow~\cite{tanaka2020pruning} aim to prune the neural networks without any training by removing the weights least salient with respect to the loss for few images from the dataset. Even though these methods surpass the trivial baseline of random network pruning at initialization, ~\cite{frankle2020pruning} suggests that accuracy of these methods on standard classification benchmarks remains below the dense network obtained after training. 

\textbf{Pruning after training} introduced by the Lottery Ticket Hypothesis~\cite{frankle2018lottery}, adapted by various works~\cite{frankle2019stabilizing,malach2020proving,frankle2020linear,girish2021lottery,chen2021nerv,chen2020lottery,Movva2020DissectingLT,desai2019evaluating,Brix2020SuccessfullyAT,yu2020playing,you2019drawing}, prune the neural network weights based on the magnitude after the network is trained. Typically the dense network is reset to the initial weights after pruning and retrained from scratch. This results in additional training cost, however, the resulting network can achieve similar performance as the dense network, and better performance than the sparse networks obtained without training. 

The third class of methods, perform the \textbf{pruning while training} ~\cite{han2015deep,han2015learning,de2020progressive,tanaka2020pruning,verdenius2020pruning,renda2020comparing}. In these methods, the network is iteratively pruned during training based on either the magnitude of weights or their gradients. These heuristics have shown to perform well, however can only induce unstructured sparsity in the network which is not hardware friendly. Further, in these methods the pruning heuristic is often agnostic to the actual accuracy criteria for which the network is being optimized. Structured pruning methods~\cite{wen2016learning,louizos2017learning,liu2017learning,li2016pruning,he2018soft,he2019filter,luo2017thinet,zhou2019accelerate,you2019gate,zhuang2018discrimination,he2017channel} on the other hand, introduce additional priors or structure on the weights to be pruned, and have been shown to be effective in improving the inference speeds of the model on both CPUs and GPUs.

\subsection{Model Quantization}
Quantization methods for compressing neural network discretize the parameters of a network to a small, finite set of values, so that they can be further stored efficiently using one or more entropy coding methods~\cite{rissanen1981universal,huffman1952method}. Some of the earlier methods resort to uniform quantization of weights to binary or tertiary representation of the weights~\cite{courbariaux2015binaryconnect,li2016ternary,zhou2018explicit,zhu2016trained,rastegari2016xnor,hubara2017quantized,jacob2018quantization}. Several other methods have focused on non-uniform \textbf{Scalar Quantization} (SQ) techniques~\cite{tung2018clip,zhou2017incremental,nagel2019data,banner2018post,wu2016quantized,zhang2018lq}. In SQ methods, neural network parameters or parameter groups can be quantized, both uniformly or non-uniformly, before or afer training. The maximum possible set of values that the representers (of the neural network parameters) can take is given by Kronecker product of representable scalar elements. 

\textbf{Vector Quantization} (VQ)~\cite{gong2014compressing,stock2019and,wang2016cnnpack,chen2015compressing,chen2016compressing} on the other hand, is a more general technique, where the representers can take any value. VQ can be done by clustering of CNN layers at various rate-accuracy trade-offs~\cite{faraone2018syq,son2018clustering}, hashing~\cite{chen2015compressing,chen2016compressing}, or residual quantization~\cite{gong2014compressing}. In practice, Scalar Quantization (SQ) is preferred over VQ due to its simplicity and effectiveness without additional encoding complexity. Young \etal~\cite{young2021transform} provide an overview of various transform based quantization techniques for compressing CNNs.

Oktay \etal~\cite{oktay2019scalable} propose to warp the network parameters in a `latent' space using a transformation, and perform scalar quantization in the learned latent space during model training. This allows SQ to be more flexible while at the same time, computationally more feasible as compared to VQ. This approach while effective at reducing the model size on the disk, doesn't provide any computation gains during inference since the decoded model is dense. In this paper, we build upon \cite{oktay2019scalable}, to jointly train a model for both compression and computational gains. We describe our approach in \cref{sec:approach}, while highlighting our contributions. We show the benefits of our approach in terms of even smaller model size and better computational gains while maintaining high levels of accuracy in \cref{sec:experiments}.

\section{Approach}
\label{sec:approach}

\begin{figure}[!t]
\centering
\includegraphics[width=0.9\linewidth]{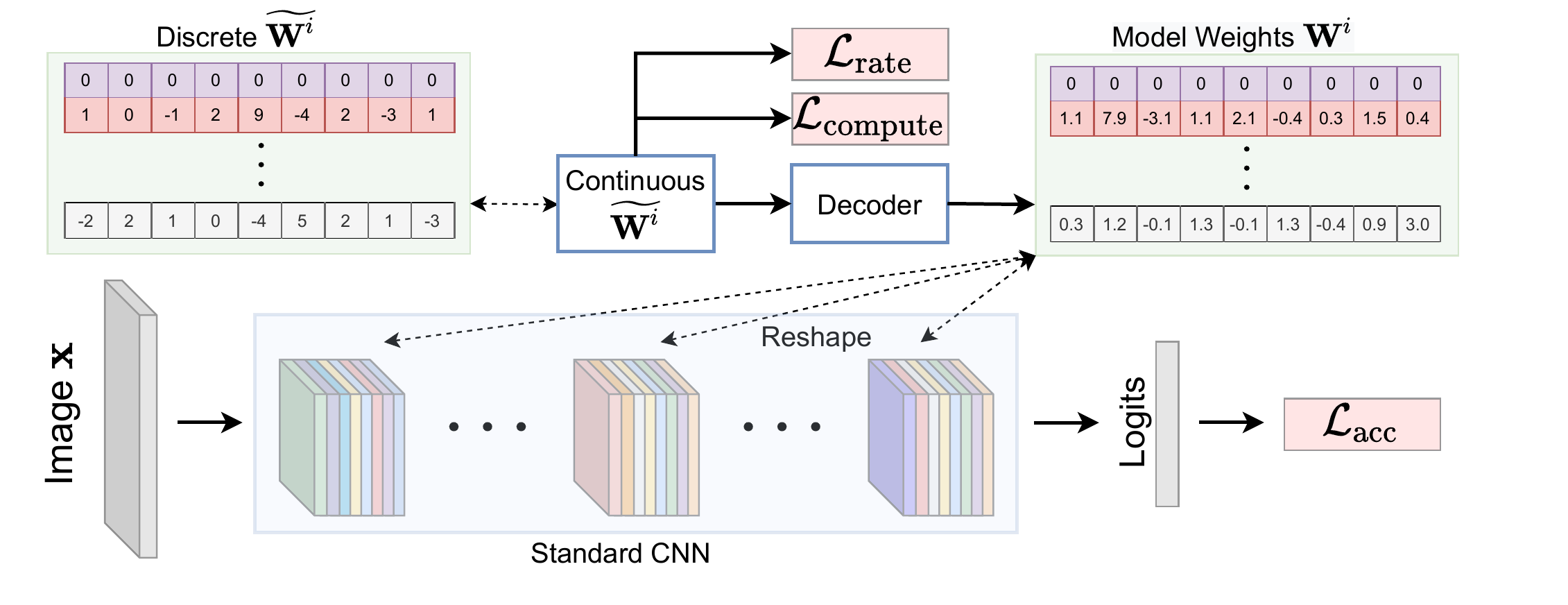}
\caption{\textbf{Overview of our approach.} A standard CNN comprises of a sequence of convolutional and fully-connected layers. We reparameterize the parameters $\bw^i$ of each of these layers as $\bwt^i$ in a quantized latent space. CNN parameters can be computed using a learned purely linear transform $\bpsi$ of the latent parameters. Linearity of transform allows sparsity in the quantized latents to translate into the sparsity of network parameters. Further, we organize each parameter tensor as a set of slices (depicted as colored bands) corresponding to different channels. Additional loss terms are then introduced to encourage slice sparsity and jointly optimize for accuracy-rate-computation.}
\vspace{-1.5em}
\label{fig:arch}
\end{figure}

We consider the task of classification using a convolutional neural network (CNN), although our approach can be trivially extended to other tasks such as object detection or generative modeling. Given a dataset of $N$ images and their corresponding labels $\{\mathbf{x}_i, y_i\}_{i=1}^{N}$, our goal is to train a CNN with parameters $\btheta$ that is jointly optimized to: 1) maximize classification accuracy, 2) minimize the number of bits required to store the model on disk, and 3) minimize the computational cost of inference in the model. 
To keep our method end-to-end trainable, we formulate it as minimization of a joint objective that allows for an accuracy-rate-computation trade-off as below
\begin{equation}
    \mathcal L(\btheta) = \mathcal L_{\text{acc}}(\btheta) + \mathcal L_{\text{rate}}(\btheta) + \mathcal L_{\text{compute}}(\btheta).
\end{equation}
For the task of classification, the accuracy term $\mathcal L_{\text{acc}}(\btheta)$ is the usual cross-entropy loss. It encourages the model to maximize prediction accuracy. The rate term $\mathcal L_{\text{rate}}(\btheta)$ encourages the model have a small disk size. We modify Oktay \etal~\cite{oktay2019scalable} in the formulation of this term, where a self-information penalty encourages the parameters to have a small bit representation. The rate term, while encouraging smaller model size, doesn't lead to computational gains as the learned parameters are still dense. Our computation term $\mathcal L_{\text{compute}}(\btheta)$ addresses this issue by introducing a structured sparsity inducing term that encourages the model to be more amenable to computational optimizations. Refer to \Cref{fig:arch} for a high-level overview of our approach. In the following sections we describe the rate and computation terms in more detail.

\subsection{Rate term}
\label{ssec:weight_reparam}
We formulate our rate term by building upon Oktay \etal~\cite{oktay2019scalable}.%
The CNN weights $\btheta$ are reparameterized in terms of a latent representation. The latent representation is quantized and compressed while the network parameters are implicitly defined as a transform of the latent representation.

A typical CNN consists of a sequence of convolutional and dense layers. For each of these layers, we denote weights %
as $\bw$ and bias as $\bb$,
\begin{align}
\btheta = \left\{\bw^1,\bb^1,\bw^2,\bb^2,\dots,\bw^N,\bb^N\right\}
\end{align}

where $N$ is the total number of layers in the network, and 
$\bw^k,\bb^k$ represent weight and bias parameters of the $k$th layer. Each of these parameters can take continuous values during inference. However these parameters are stored using their discrete reparameterized forms belonging to a corresponding set
\begin{align}
\bphi = \left\{\bwt^1,\bbt^1,\bwt^2,\bbt^2,\dots,\bwt^N,\bbt^N\right\}
\end{align}

For each convolutional layer, $\bw$ is a weight tensor of dimensions $ \cin \times \cout \times K\times K$, where $\cin$ is the number of input channels, and $\cout$ is the number of output channels, and $K$ denotes the filter width and height. The corresponding reparameterized weight tensor $\bwt$ is represented by a two-dimensional (2D) matrix of size $\cin \cout \times K^2 $. Similarly, for each dense layer, $\bw$ is a tensor of dimension $\cin \times \cout$ and its corresponding reparameterized weight $\bwt$ can be represented by a matrix of dimension $\cin \cout \times 1$. All the biases can be reparameterized in the same way as dense layers. Each reparameterization can hence be represented by a quantized 2D matrix $\bwt \in \mathbb{Z}^{\cin\cout \times l}$ where $l=1$ for dense weights (and biases) while $l=K^2$ for convolutional weights. Each in $\bwt$ represents a sample drawn from an $l$-dimensional discrete probability distribution. In order to convert parameters from latent space $\bphi$ to model space $\btheta$, \cite{oktay2019scalable} introduced learnable affine transform $\bpsi$ for convolutional, dense, and bias weights. 
\begin{align}
\bw = \text{reshape} (\bpsiscale (\bwt^T + \bpsishift))
\label{eq:affine}
\end{align}

where $\bpsiscale \in \mathbb{R}^{l\times l},\bpsishift \in \mathbb{R}^{l}$ are the affine transformation parameters. Also note that model partitioning is performed such that different kinds of layers use different pair of transform parameters $(\bpsiscale, \bpsishift)$. That is, convolutional layers have their own transform, while dense layers have their own, and so on.

As $\bphi$ consists of discrete parameters which are difficult to optimize, continuous surrogates $\widehat{\bw}$ are maintained for each quantized parameter $\bwt$. $\widehat{\bw}$ is thus simply obtained by rounding the individual elements of $\bwt$ to the nearest integer. Rate minimization is achieved by enforcing an entropy penalty on the discrete weights. Since discrete entropy is non-differentiable, a continuous proxy is used instead. A set of continuous density functions implemented as small neural networks are fitted to $\widehat{\bw}+n$ where $n\sim\mathcal{U}(-\frac{1}{2},\frac{1}{2})$ is a matrix whose elements are drawn from the uniform distribution. We recap that each latent weight layer $\bwt$ is a matrix in $\mathbb{Z}^{\cin\cout \times l}$ and can be rewritten as a factorized probability model as shown in \cref{eq:pmodel}. The entropy of model weights can now be minimized directly by minimizing the negative log likelihood which serve as an approximation to the self-information $I$ (\cref{eq:pmodel}). The rate term is then the sum of all the self-information terms.
\begin{align}
    q(\bwt) = \prod_{j=1}^{\cin \cout} \prod_{i=1}^{l} q_i(\bwt_{j, i}) &\quad\text{and}\quad  I(\bwt) = -\log_2 {q(\bwt)} \label{eq:pmodel} \\
    \mathcal L_{\text{rate}}(\btheta) &= \lambdai \sum_{\bsphi\in \bphi} I(\bsphi)
\end{align}
Here $q_i$ represents the probability model for the $i^\text{th}$ element of an $l-$dimensional row slice of $\bwt$, $\lambdai$ is a trade-off parameter that permits specification of relative weight with respect to other terms.
A straight-through estimator~\cite{bengio2013estimating} is used to back-propagate the gradients from the classification loss to parameter latents. 

\subsection{Sparsity priors}

While the rate term described in previous section encourages a smaller representation of the model in terms of the number of bits required to store the model on disk, it doesn't reduce the number of parameters in a way that leads to computational gains. To address this, we introduce a few key changes and then formulate our computation term as structured sparsity priors that lead to reduced computation. Note that we formulate all the priors in the reparameterized latent space to decouple from the affine transform parameters $\bpsiscale, \bpsishift$ and to be consistent with the $\mathcal L_{\text{rate}}(\btheta)$ term that is also applied in the same space.

\smallskip
\noindent
\textbf{Unstructured sparsity with zero mean prior:}
Our first contribution derives from the observation that even if the latent representation of the parameters is all zero, the resulting parameters may still be non-zero. For example, in order to enforce structural sparsity in the convolutional layers in the model space $\bw$, we require each $K\times K$ slice to be all zero. However, this is only possible if $(\bpsishift+\bwt)$ in Eq.~\ref{eq:affine} is a zero vector itself or lies in the null space of $\bpsiscale$. We notice that the latter does not occur in most practical situations especially when the vector is discrete. Therefore, we remove the shift parameter $\bpsishift$ and make the affine transform a purely linear transform. Now a zero vector in latent space results in a zero vector in the model space. Note that any single non zero element, in the latents corresponding to a slice, causes the full transformed vector to be nonzero and does not yield any sparsity in the model space.

Next, we observe that zero entropy distributions over latents do not necessarily imply zero valued latents. They could take on any constant value. Therefore, we introduce a zero-mean prior on the latent representation of the parameters to encourage the latent parameters to go to zero as the entropy of their distribution decreases. This penalty encourages the ground state of the zero entropy distributions to be zero value, as opposed to an arbitrary constant values in absence of this penalty. As a result this penalty encourages unstructured sparsity in the latents. We use a simple gaussian as a zero mean prior that results in an $l_2$ penalty on the continuous surrogates $\boldsymbol{\widehat{\bw}}$ of the individual weight latents. A laplacian prior, resulting in a $l_1$ penalty, can also be used along with/instead of the gaussian prior. While $l_1$ penalty usually leads to sparsity on its own, in conjunction with rate penalty both seemed to have similar effect of encouraging unstructured sparsity. We experimented with both the priors and observed similar performance (an ablation study is provided in Sec. \ref{supp_sec:ablation_norm} of the appendix).

\smallskip
\noindent
\textbf{Group sparsity prior for computation:}
Now note that $\bwt$ can be viewed as being represented as a set of groups (or slices), where each group is of size $K \times K$ corresponding to a same size slice of $\bw$. These slices going to zero as a whole lead to a block sparse structure in the corresponding transform matrix for the parameter $\bw$, which can then be exploited for computational gains. To encourage individual slices to go to zero as a whole we propose to use a group sparsity regularization~\cite{yuan2006model} on the $\widehat{\bw}$ slices as following
\begin{align}
    \mathcal{L}_\text{group} = \sum_{j=1}^{\cin \times \cout} \sqrt{\rho_j} \|\widehat{\bw}_j\|_2.
\end{align}
Here $\bwt_j$ is the $j$th slice of the latent weight matrix, and $\rho_j$ is a term to account for varying group sizes. Our overall computation term based on structured and unstructured sparsity becomes
\begin{align}
    \mathcal L_{\text{compute}}(\btheta) = \lambdau \|\widehat{\bw}\|^2_2 + \lambdas\sum_{j=1}^{\cin \times \cout} \sqrt{\rho_j}\|\widehat{\bw}_j\|_2
\end{align}
Note that $\lambdau$ and $\lambdas$ are trade-off parameters for the individual sparsity terms. 

\subsection{Discussion}
\label{ssec:discussion}
The overall loss function is the combination of cross-entropy loss (for classification), self-information of the reparameterizations, and the regularization for structured and unstructured sparsity as following

{\small
\begin{align}
    \underbrace{\sum_{(\mathbf{x}, y) \sim D} - \log{p(y | \mathbf{x}; \bwt)}}_{\text{Cross Entropy}} + \underbrace{\vphantom{ \sum_{(\mathbf{x}, y) \sim D} } \lambdai \sum_{\bsphi\in \bphi} I(\bsphi)}_{\substack{\text{Parameter} \\ \text{Entropy}}} %
+  \underbrace{\vphantom{\sum_{j=1}^{\cin \times \cout}}\lambdau \|\widehat{\bw}\|^2_2}_{\substack{\text{Unstructured} \\ \text{Sparsity}}}+ 
    \underbrace{\lambdas\sum_{j=1}^{\cin \times \cout} \sqrt{\rho_j}\|\widehat{\bw}_j\|_2}_{\substack{\text{Structured} \\ \text{Sparsity}}}  \label{eq:loss}
\end{align}
}%

Without the group prior, the gaussian prior and the learned prior used by the rate penalty assume the individual weights in each slice to be independent draws. The group prior enforces structure on the weights and assigns higher likelihood for fully sparse or dense weight slices. Combined, all these priors enforce a slice-wise sparsity in the decoded weights. This sparsity, which we term slice-sparsity, hits a middle ground between completely unstructured sparsity where individual weights are independently zero and fully structured sparsity where an entire filter is zero. Note that convolutions with slice-sparse weight tensors, when represented as matrix multiplications (as done by the popular im2col algorithm) \cite{chellapilla2006high}, form block sparse matrices. Libraries, such as NVIDIA's Cusparse library \cite{narang2017block}, can exploit such block sparse structure to achieve speedups in comparison to dense matrix multiplications or entirely unstructured sparsity. We use the term Slice FLOPs (or SFLOPs) to denote the direct computational gain we can obtain through block or slice sparsity.

\section{Experiments}
\label{sec:experiments}

\medskip
\noindent
\textbf{Datasets.} We consider three datasets in our experiments. CIFAR-10 and CIFAR-100 datasets~\cite{krizhevsky2009learning} consist of 50000 training and 10000 test color images each of size $32\times32$. CIFAR-10 has 10 classes (with 6000 images per class) and CIFAR-100 has 100 classes (with 600 images per class). For large scale experiments, we use ILSVRC2012 (ImageNet) dataset~\cite{deng2009imagenet}. It has 1.2 million images for training, 50000 images for the test and 1000 classes. 

\medskip
\noindent
\textbf{Network Architectures.}
For CIFAR-10 and CIFAR-100 datasets, we show results using two architectures - VGG-16~\cite{simonyan2014very} and ResNet-20 with a width multiplier of 4 (ResNet-20-4)~\cite{he2016deep}. VGG-16 is a commonly used architecture consisting of 13 convolutional layers of kernel size $3\times 3$ and 3 dense or fully-connected layers. Dense layers are resized to adapt to CIFAR's $32\times32$ image size, as done in the baseline approaches~\cite{oktay2019scalable}. ResNet-20-4 consists of 3 ResNet groups, each with 3 residual blocks. All the convolutional layer are of size $3\times 3$, and there is a single dense layer at the end. 
For the ImageNet experiments, we use ResNet-18/50 \cite{he2016deep}, an 18/50-layer network comprising of one $7\times7$ convolutional layer, one dense layer and the remaining $3\times3$ convolutional layers. We also run experiments on the MobileNet-V2 architecture \cite{sandler2018mobilenetv2} which consists of depthwise separable convolutions and inverted Bottleneck blocks.

\subsection{Implementation Details}
We train all of our models from scratch. CIFAR-10 and CIFAR-100 experiments are trained for 200 epochs. We use the FFCV library~\cite{leclerc2022ffcv} for faster ImageNet training, with a batchsize of 2048/512 for ResNet-18/50 split across 4 GPUs. We train ResNet-18/50 for 35/32 epochs in order to keep the range of the uncompressed network accuracies similar to other works for a fair comparison. CIFAR-10 and CIFAR-100 training schedules amount to much fewer iterations compared to previous works. Nevertheless, we show strong performance in terms of model compression and sparsity outperforming existing model compression works while converging faster with relatively fewer epochs.

We use two optimizers, one for optimizing the parameters of the entropy model $q(\bwt)$, and the other for optimizing the decoder matrix $\bpsi$ and the model parameters $\bphi$. Entropy model is optimized using Adam \cite{kingma2014adam} with a learning rate of $10^{-4}$ for all our experiments. The remaining parameters are optimized using Adam with a learning rate of $0.01$ for CIFAR-10 experiments and a learning rate of $0.02/0.006/0.01$ for ResNet-18/50/MobileNet-v2 respectively on ImageNet with a cosine decay schedule. Our model compression results are reported using the \textsf{torchac} library~\cite{mentzer2019practical} which does arithmetic coding of the weights given probability tables for the quantized values which we obtain from the trained entropy models. We do not compress the biases and batch normalization parameters and include the additional sizes from these parameters as well as the parameters $\boldsymbol{\Psi}$ whenever we report the model size unless mentioned otherwise. The entropy model for VGG-16 consists of a parameter group for each dense layer and a parameter group for all $3\times3$ convolutions leading to four weight decoders/probability models for each parameter group. For ResNet-20-4 we use zero padding shortcut type A as defined in \cite{he2016deep}, which leads to only 2 parameter groups, one for the final dense layer and the other for all $3\times3$ convolutions. For ResNet-18 trained on ImageNet, we use three parameter groups, for the initial 7x7 convolution, $3\times3$ convolutions, as well as the dense layer. ResNet-50 consists of an additional parameter group for $1\times1$ convolutions. MobileNet-V2 consists of 3 parameter groups for the initial 3x3 convolution, final dense layer and the remaining 3x3 convolution.
Initialization of the network weights and the decoders are non-trivial in terms of this decoding scheme. We provide an explanation of our initialization approach in Sec. \ref{supp_sec:init} in our appendix.
Recall from the ~\cref{eq:loss} that in addition to the usual cross-entropy loss for classification, we have three additional loss terms and coefficients. 

\medskip
\noindent
\textbf{Compression coefficient.} The rate term allows us to trade-off between model size (bit-rate) and accuracy. We fix the coefficient $\lambdai=10^{-4}$ in our experiments.

\medskip
\noindent
\textbf{Sparsity coefficients.}
Note that the sparsity terms not only improve the model's inference speed (by reducing the number of FLOPs) but also reduces the entropy of latent weights $\bwt$ as most of the weights become zero.
By varying the two trade-off parameters, one for each sparsity term, we obtain different points on the pareto curves for accuracy-model size trade-off and accuracy-sparsity trade-off. We study each of these trade-offs extensively in~\cref{ssec:ablation}.

\subsection{Comparison with compression methods}
\label{ssec:quantitative}

\renewcommand{\arraystretch}{1.1}
\begin{table}[t]
\centering
\resizebox{0.48\linewidth}{!}{
\begin{tabular}{@{}L{\dimexpr.27\linewidth}R{\dimexpr.17\linewidth}R{\dimexpr.12\linewidth}R{\dimexpr.1\linewidth}@{}}
\toprule
Algorithm & \makecell{Size \\(KB)} & \makecell{Error \\(Top-1 \%)} & \makecell{Sparsity \\(\%)}\\
\midrule
\multicolumn{4}{c}{VGG-16(CIFAR-10)}\\
\midrule
Uncompressed & 60 MB (1$\times$)& 6.6 & 0 \\
BC-GHS\cite{louizos2017bayesian} & 525 (116$\times$) & 9.2 & 94.5\\
DeepCABAC~\cite{wiedemann2019deepcabac} & 960 (62$\times$) & 9.0 & 92.4\\
MRCL~\cite{havasi2018minimal} & 168 (452$\times$) & 10.0 & 0 \\
Oktay \etal \cite{oktay2019scalable} & 101 (590$\times$) & 10.0 & 0 \\
\ours (Best) & 129 (465x) & 7.4 & 97.4 \\
\ours (Extreme) & \textbf{76 (800x)} & 10 &\textbf{99.2}\\
\midrule
\multicolumn{4}{c}{ResNet-20-4(CIFAR-10)}\\
\midrule
Uncompressed & 17.2 MB (1$\times$) & 5.5 & 0 \\
Oktay \etal \cite{oktay2019scalable} & 128 (134$\times$) & 8.8& 0\\
\ours (Best) & 139 (123$\times$) & 6&88.5\\
\ours (Extreme) & \textbf{66 (282$\times$)} & 8.5&\textbf{97.9}\\
\midrule
\multicolumn{4}{c}{ResNet-20-4(CIFAR-100)}\\
\midrule
Uncompressed & 17.2 MB (1$\times$) & 26.8  & 0 \\
\ours (Best) & 125 (137$\times$) & 27.0&86.6\\
\ours (Extreme) & \textbf{76 (226$\times$)} & 30.9&\textbf{97.1}\\
\bottomrule
\end{tabular}
} %
\quad
\resizebox{0.48\linewidth}{!}{

\begin{tabular}{@{}L{\dimexpr.27\linewidth}R{\dimexpr.18\linewidth}R{\dimexpr.12\linewidth}R{\dimexpr.1\linewidth}@{}}
\toprule
Algorithm & \makecell{Size \\(MB)} & \makecell{Error \\(Top-1 \%)} & \makecell{Sparsity \\(\%)}\\
\midrule
\multicolumn{4}{c}{ResNet-18 (ImageNet)}\\
\midrule
Uncompressed & 46.7 (1$\times$) & 30.0&0\\
AP + Coreset-S~\cite{dubey2018coreset} & 3.11 (15$\times$) & 32 & -\\
Choi \etal~\cite{choi2018compressionshort}& 1.93 ($24\times$) & 32.6&80$^*$\\
Oktay \etal \cite{oktay2019scalable} & 1.97 (24$\times$) & 30.0 & 0 \\

\ours (Best) & 1.01 (46$\times$) & 31.2&58.3\\
\ours (Extreme) & \textbf{0.86} (54$\times$) & 32.3&\textbf{65.2}\\
\midrule
\multicolumn{4}{c}{ResNet-50 (ImageNet)}\\
\midrule
Uncompressed & 102 (1$\times$) & 23.4&0\\
Lin \etal~\cite{lin2019toward}&39.91 (3$\times$) & 27.7& 39.2\\
Frankle \etal \cite{frankle2019stabilizing} & 20.4 (5$\times$) & 24.0&49.1\\
AP + Coreset-S~\cite{dubey2018coreset} & 6.46 (16$\times$) & 26.0 & -\\
DeepCABAC~\cite{wiedemann2019deepcabac}&6.06 (17$\times$) & 25.9& 25.4\\
Oktay \etal \cite{oktay2019scalable} & 5.49 (19$\times$) & 26.0& 0 \\
\ours(Best) & 3.60 (29$\times$) & 26.3&42.1\\
\ours(Extreme) & \textbf{3.00 (34$\times$)} & 26.9&\textbf{51.6}\\

\midrule
\multicolumn{4}{c}{MobileNet-V2 (ImageNet)}\\
\midrule
Uncompressed & 14.0 (1$\times$) & 32.7& 0.0\\
Tu \etal\cite{tu2020pruning}& 10.3 (1.3$\times$) & 32.7 & 25.0\\
Wang \etal\cite{wang2019haq}& 0.95 (15$\times$) & 33.5 & 0\\
LilNetX (Best)& 0.67 (21$\times$) & 32.8 & 56.8\\
LilNetX (Extreme)& \textbf{0.56 (25$\times$)} & 33.3& \textbf{64.7}\\
\bottomrule
\end{tabular}
} %
\caption{Comparison of our approach against other model compression techniques. We show two cases of our method with best error rate and also extreme compression at higher error rate. We achieve higher compression along with the added computational benefits of high slice sparsity. Best corresponds to our best model in terms of accuracy while Extreme is matching the range of error of baselines if exists. $^*$ corresponds to unstructured sparsity which is generally higher than structured sparsity. $-$ implies that the work perform pruning but do not report numbers in their paper.
}
\vspace{-1em}
\label{tab:baseline_comparison}

\end{table}

As discussed in~\Cref{sec:related}, existing approaches for model compression follow either \textbf{quantization}, \textbf{pruning}, or both. We compare with the state of the art methods in each of these two categories. Among model quantization methods, we use Oktay \etal~\cite{oktay2019scalable} and Dubey \etal~\cite{dubey2018coreset} for comparison. Note that while \cite{oktay2019scalable}'s method compresses the model for storage and transmission, the uncompressed model is dense and doesn't provide any computational speedups. \cite{dubey2018coreset} prunes the filters along with quantization and helps with speed-up as well. 
Our results are summarized in Table \ref{tab:baseline_comparison}. Unless otherwise noted, we use the numbers reported by the original papers.
Since we do not have access to many prior art models, we compare using overall sparsity.
 For the CIFAR-10 dataset, we achieve the best results in compression while also achieving a lower Top-1 error rate for both VGG-16 and ResNet-20-4. For VGG-16 we obtain the best performance in the error range of $\sim7\%$ at $129KB$ which is a 465x compression compared to the baseline model. At the $\sim10\%$ error range, we outperform \cite{oktay2019scalable} in terms of model compression and also with a $99.2\%$ slice sparsity. For ResNet-20-4, compared to \cite{oktay2019scalable} we achieve almost 3 times the compression rate at a similar error rate of $\sim10.1$, while simultaneously achieving extremely high levels of slice sparsity ($98.73\%$) in the network weights.
Similar results hold for the case of CIFAR-100 where we achieve a 137$\times$ compression in model size with 86.67\% sparsity with little to no drop in accuracy compared to the uncompressed model. 

For ResNet-18 trained on ImageNet, we achieve 46$\times$ compression as compared to the uncompressed model with a small drop in accuracy. The compressed network achieves a group sparsity of 58.3\%. For an extreme compression case, we achieve higher levels of compression ($54\times$) and sparsity ($\sim65\%$) at the cost of $\sim2\%$ accuracy compared to uncompressed model. While Choi \etal achieve a sparsity of 80\%, it is unstructured and does not offer computational gains without hardware modifications. Finally, for ResNet-50, our best model achieves a compression rate of $29\times$, compared to the next best work of Oktay \etal~\cite{oktay2019scalable} with a rate of $19\times$ at a similar error rate. We additionally obtain $42.1\%$ slice structured sparsity. An extreme case of our model achieves a higher compression rate of $34\times$ with sparsity of $51.6\%$ albeit with an error rate of $26.9\%$.

To show the generality of our approach, we additionally show experiments using the MobileNetV2 \cite{sandler2018mobilenetv2} architecture trained on ImageNet. Our approach achieves almost $21\times$ compression with respect to the uncompressed model in terms of model size and achieves a sparsity of 56.8\% with almost no drop in Top-1 error on ImageNet (32.8\%). A more extreme variant, with higher sparsity constraints achieves $25\times$ compression with a sparsity of $64.7\%$. The results show that our approach works well for different types of architectures such as MobileNetV2 which consist of Inverted BottleNeck layers coupled with Depthwise Separable Convolutions. Additionally, even though MobileNet is already optimized towards size (number of parameters) and computation (FLOPs), we are able to obtain high levels of additional compression and computation reduction with little to no loss in top-1 accuracy highighting the efficacy of our approach in different settings.

We therefore conclude that the \ours~framework outperforms state-of-the-art approaches in model compression by a significant margin while also achieving  sparsification of the network weights for computational gains.

\subsection{FLOPs/Speedup-Accuracy Tradeoff}
\begin{figure}[t]
\centering
\includegraphics[width=0.015\linewidth, trim=0 -3cm 0 0, clip]{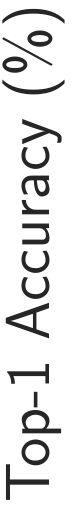}
\includegraphics[width=0.32\linewidth]{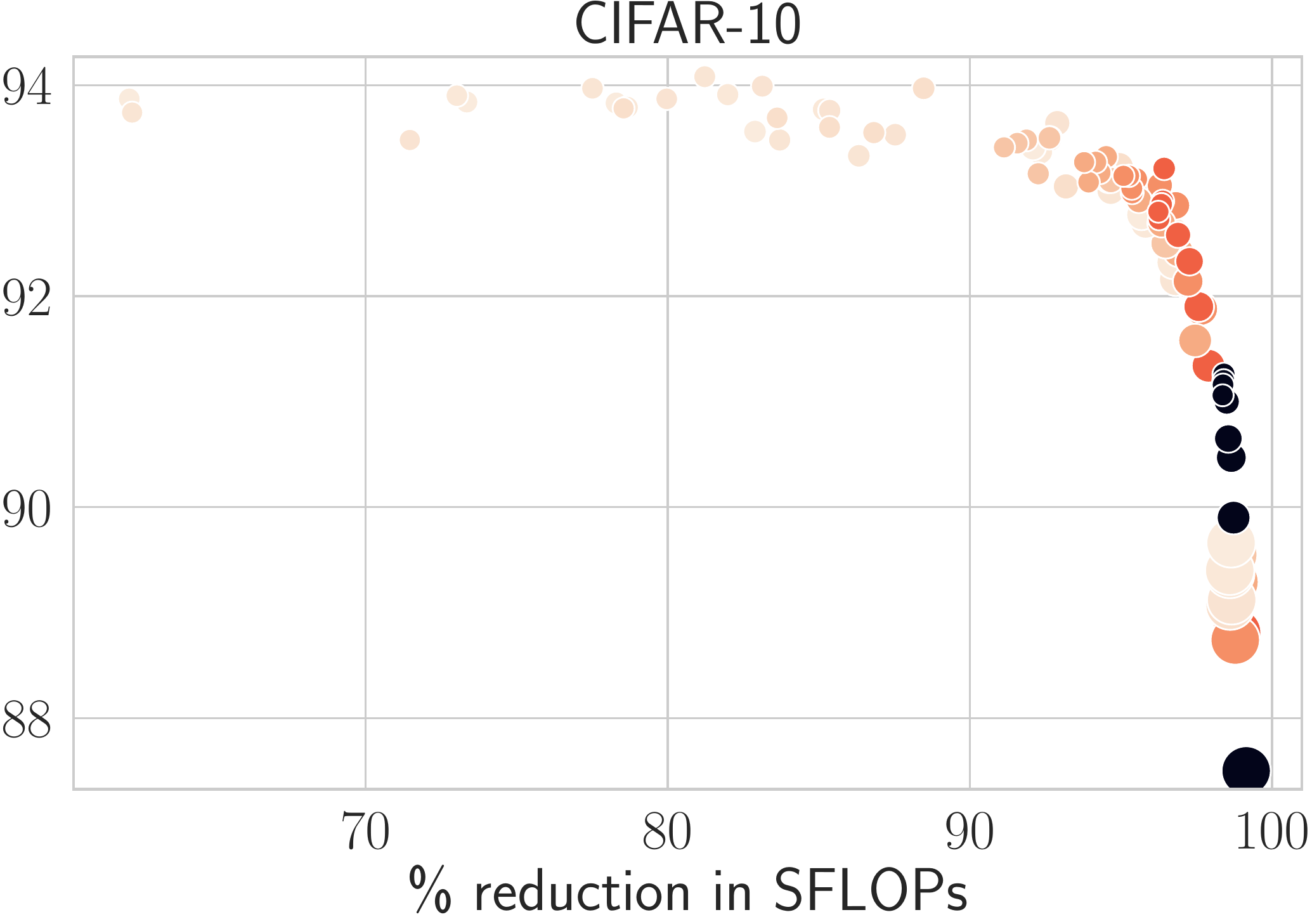}
\includegraphics[width=0.32\linewidth]{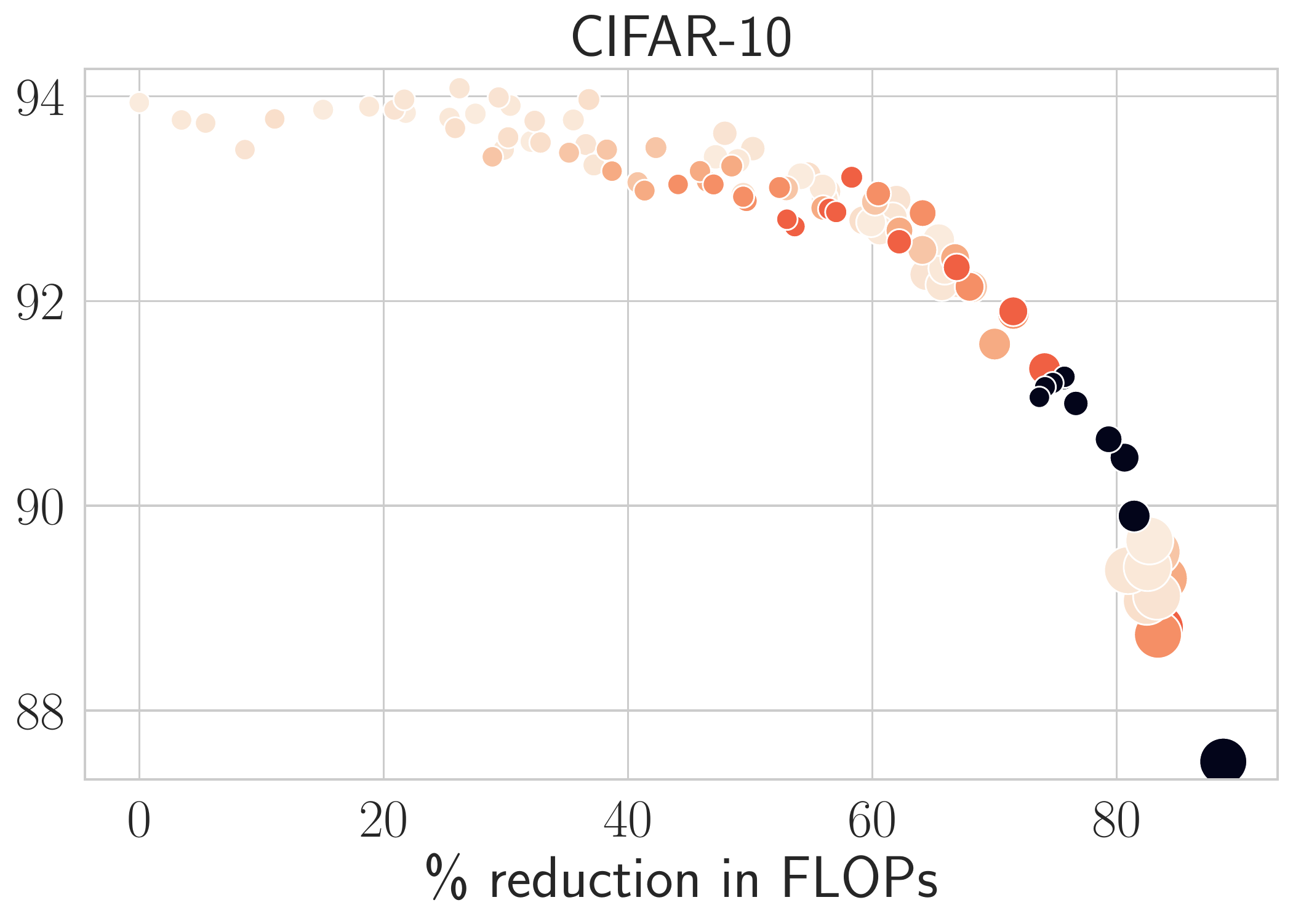}
\includegraphics[width=0.32\linewidth]{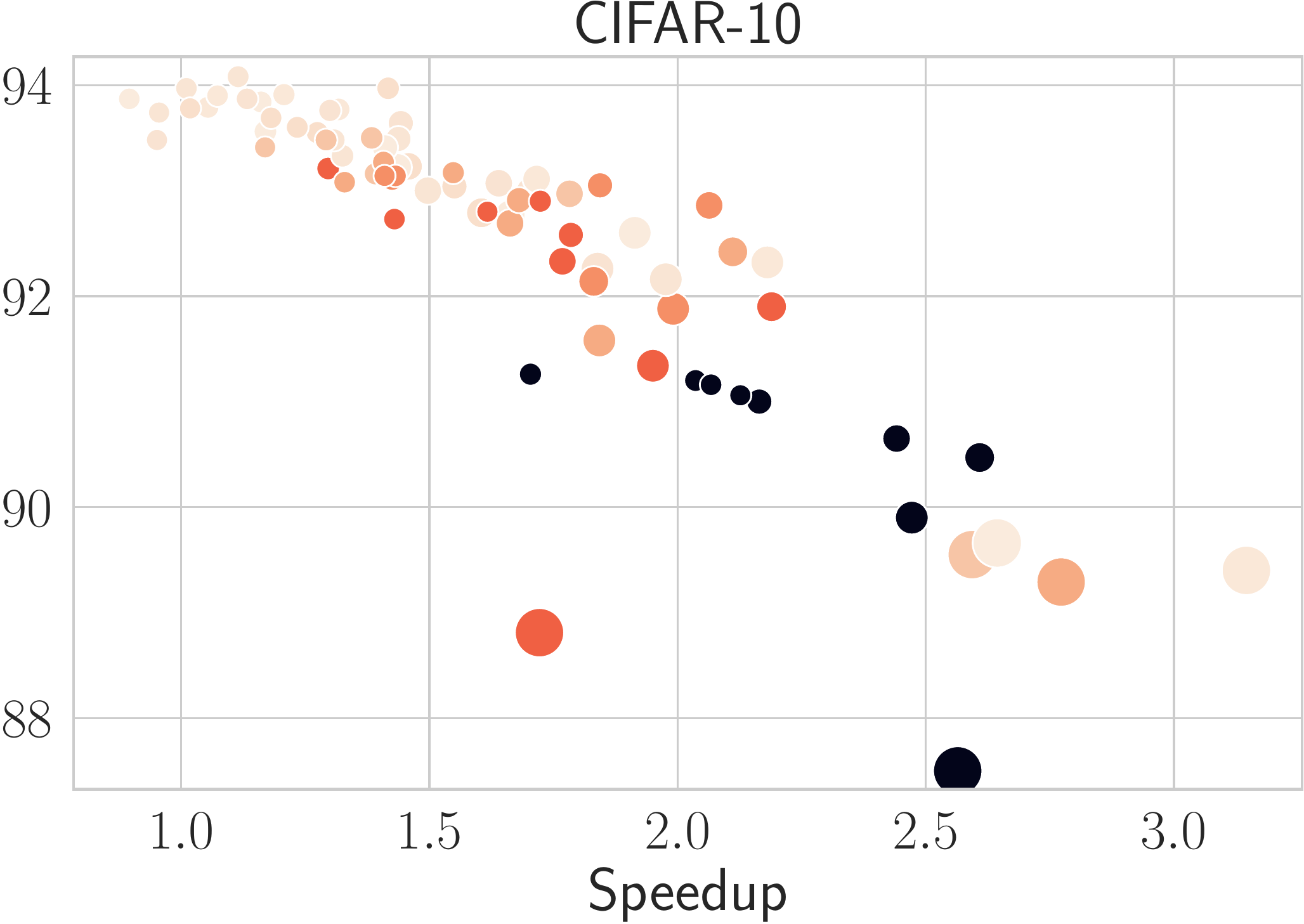}
\caption{Top-1 Accuracy \vs the $\%$ reduction in SFLOPs (left)/FLOPs (middle)/Speedup (Right) as defined in \S~\ref{ssec:discussion}. Speedup is the ratio of CPU inference times for the uncompressed to our compressed models on the CIFAR-10 test set. Hue goes from light to dark as we increase the $\lambdas$, while the marker size increases with $\lambdau$. Note that SFLOPs have a higher reduction due to higher slice sparsity compared to full structure sparsity while still offering computational gains. We obtain high $\%$ reduction in SFLOPs and FLOPs coupled with inference speedups with little loss in accuracy.
}
\label{fig:flops}
\vspace{-1em}
\end{figure}

While we show compression gains with respect to state-of-the-art in Sec.\ref{ssec:quantitative}, here we highlight the computational gains we get through slice sparsity. As explained in Sec. \ref{ssec:discussion}, slice sparsity enables us to perform block sparse multiplications which offer network speedups. We use SFLOPs to denote the theoretical FLOPs obtained by computing using only non zero weight slices. Additionally we show the FLOPs corresponding to the fully structured regime by removing entire filters or input channels of a weight tensor which are completely zeros. While not directly optimizing for FLOPs, a high slice sparsity also provides the added benefits of full structured sparsity. Through structured sparsity, we also obtain computational gains in terms of inference speedups for the compressed network. Thus, both types of sparsity can offer computational speedups with no hardware modifications. We show the tradeoff between SFLOPs, FLOPs and speedups against Top-1 Accuracy in Figure \ref{fig:flops}. Speedups are measured on CPU by timing the forward pass of the full CIFAR-10 test set with a batch size of 128. The ratio of time taken between the uncompressed model to the compressed model provides the speedup gains reported in Figure \ref{fig:flops}. Note the high \% reduction in SFLOPs compared to FLOPs due to allowing for less constrained sparsity while still offering computational gains. We see that we obtain high SFLOPs reduction ($\sim$90\%) with modest accuracy drops while also getting high levels of FLOP reduction. We additionally show the speedups in inference time for the compressed network by removing entire filters or input channels for convolutional layers based on their structural sparsity. We see from Fig. \ref{fig:flops} that we obtain almost a $1.5\times$ speedup compared to uncompressed network with no drop in Top-1 accuracy. Even higher values of speedup at $2\times$ are obtained for modest drops in accuracy $\sim 1\%$. Thus, our framework allows for computational gains directly through slice sparsity (SFLOPs) or fully structured sparsity (FLOPs). 

\section{Analysis}
\label{ssec:ablation}

In this section, we analyze the effect of the different priors in detail. We use the ResNet-20-4 architecture trained on CIFAR-10/100 datasets. We use a constant compression coefficient $\lambda_I=10^{-4}$. Since both the unstructured sparsity and the structured sparsity coefficients $\lambdau$, and $\lambdas$ impact overall sparsity, we keep one of them fixed, and analyze the impact of varying the other on model performance (accuracy), model size (bit-rate), and the slice sparsity (fraction). As the size of the weight decoders and the uncompressed parameters is fixed, we analyze model sizes for the compressed parameters in this section. 

\begin{figure*}[!t]
    \centering
\setlength{\tabcolsep}{1pt}
\begin{tabular}{ccc}
\includegraphics[width=0.025\linewidth, trim=0 -2cm 0 0, clip]{plots/ylabel.pdf}&
\includegraphics[width=0.352\linewidth]{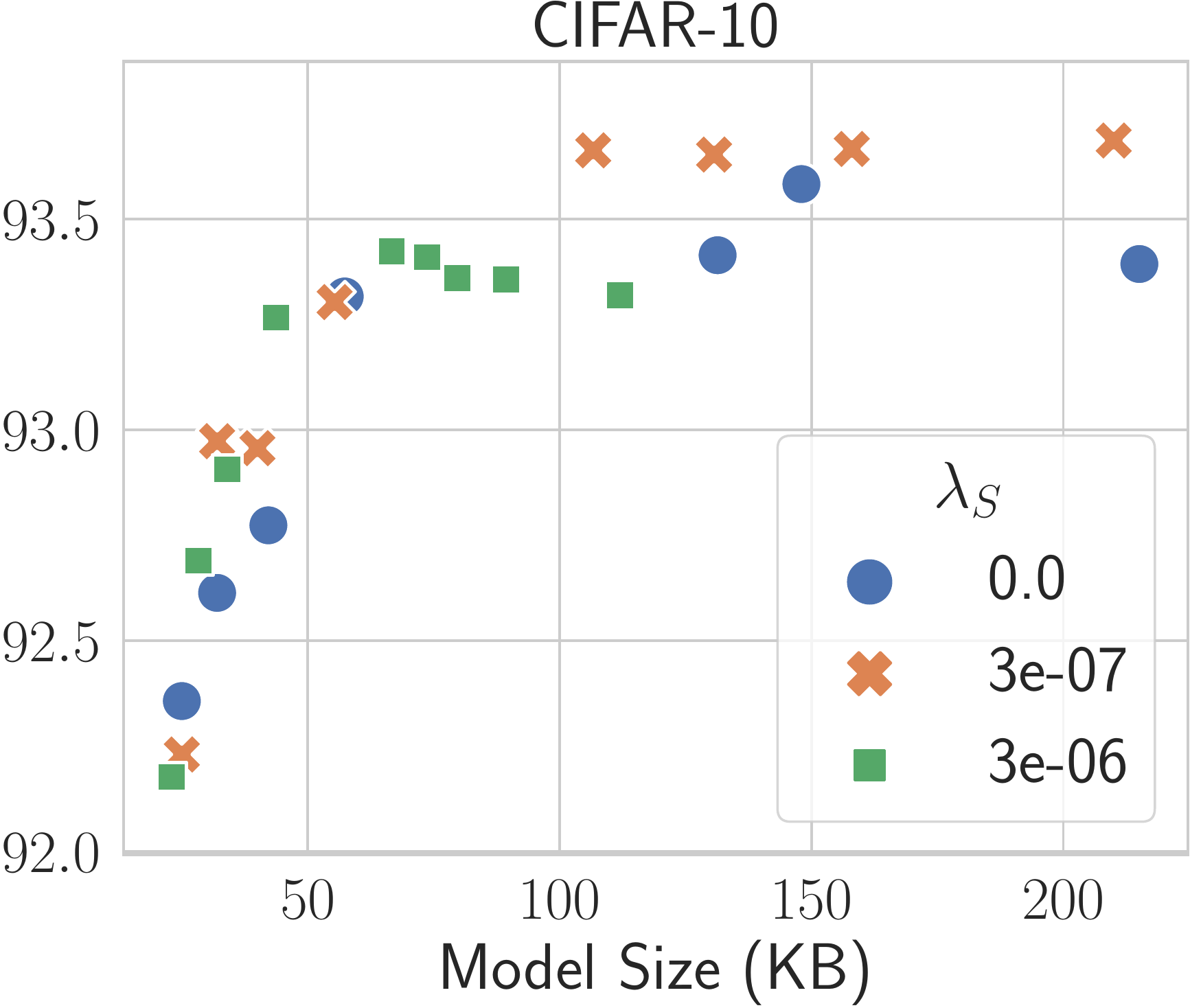}&
\includegraphics[width=0.36\linewidth]{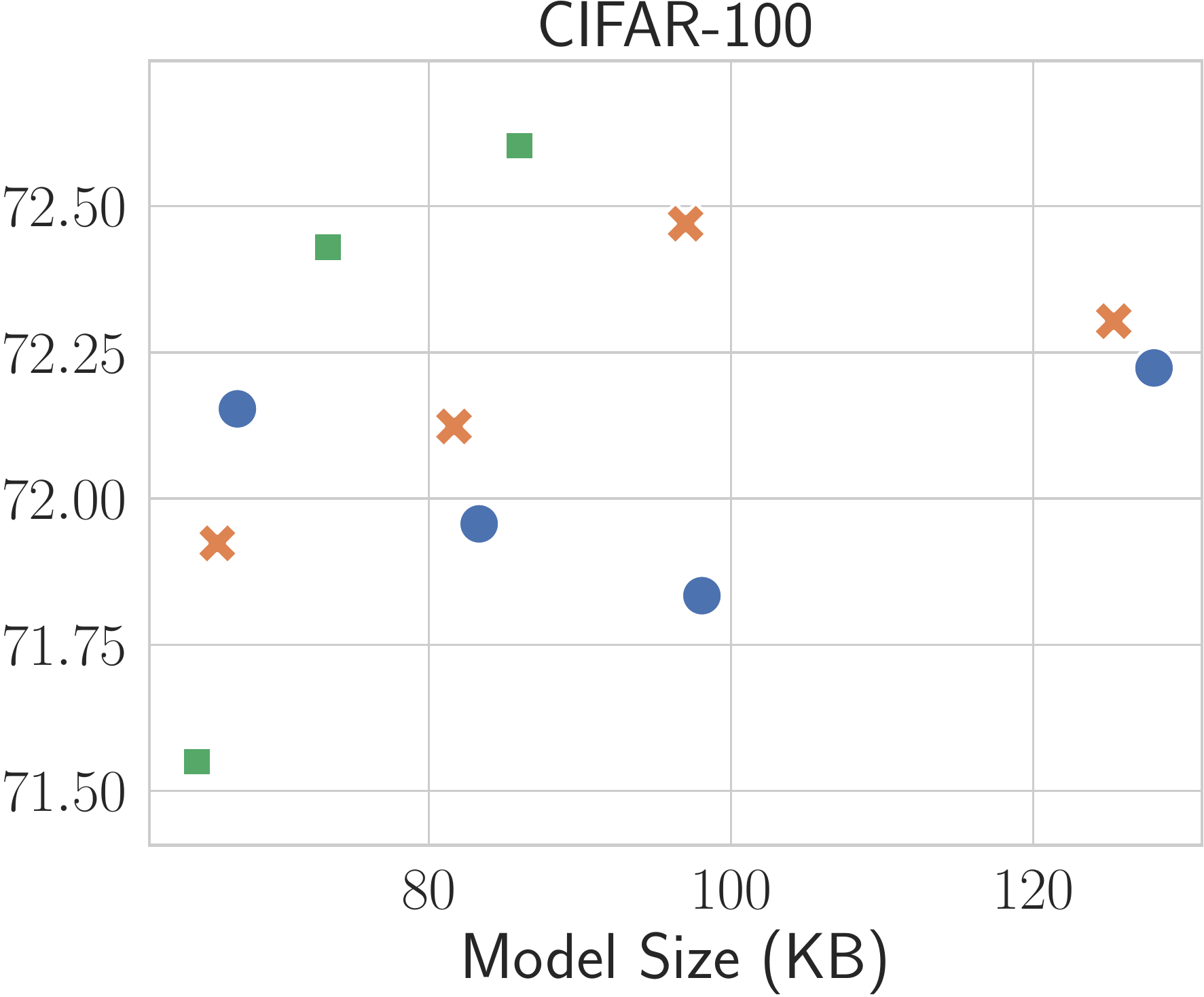}\\
&\quad\quad(a)&\quad\quad(b)\\\\
\includegraphics[width=0.025\linewidth, trim=0 -2cm 0 0, clip]{plots/ylabel.pdf}&
\includegraphics[width=0.352\linewidth]{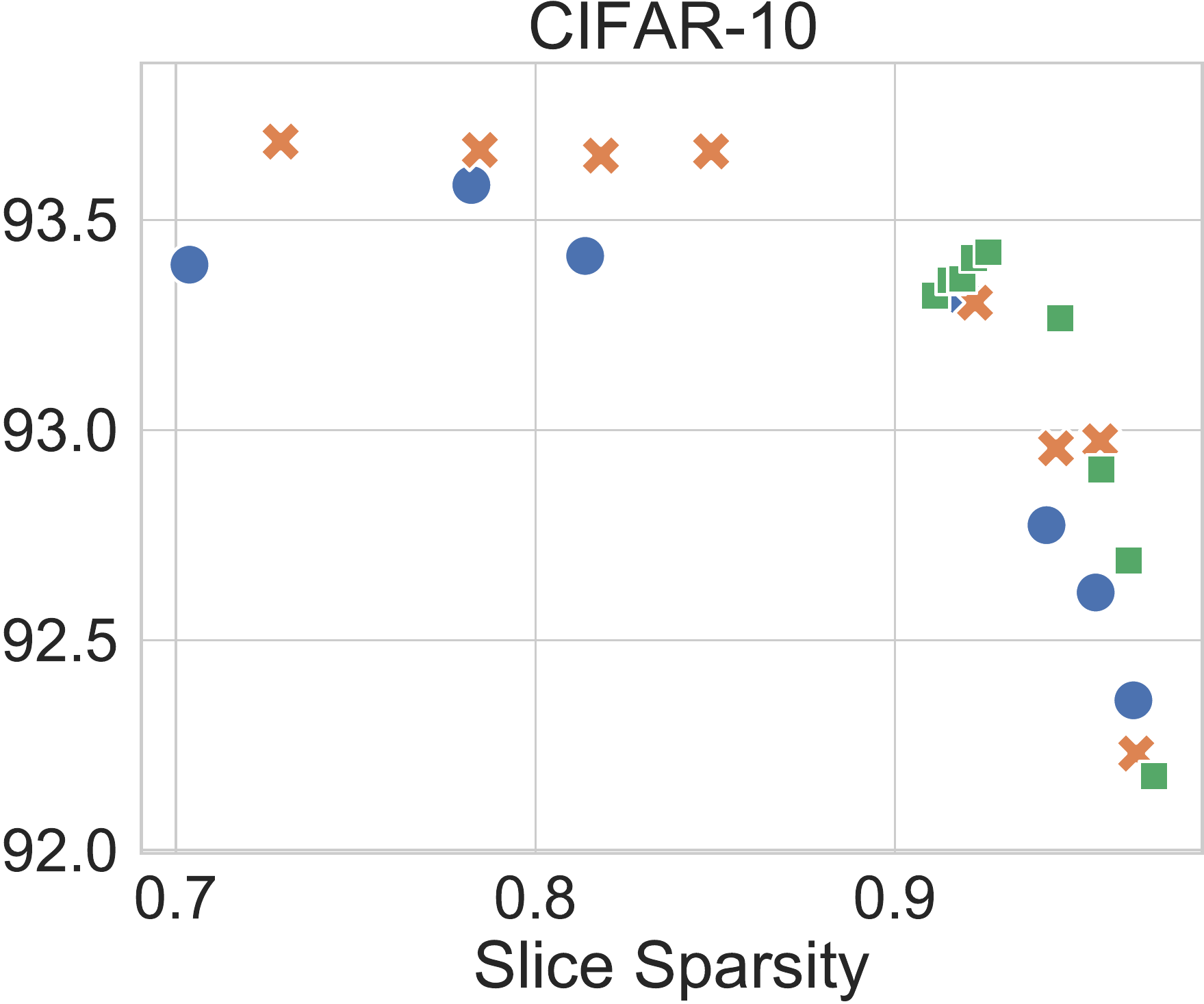}&
\includegraphics[width=0.36\linewidth]{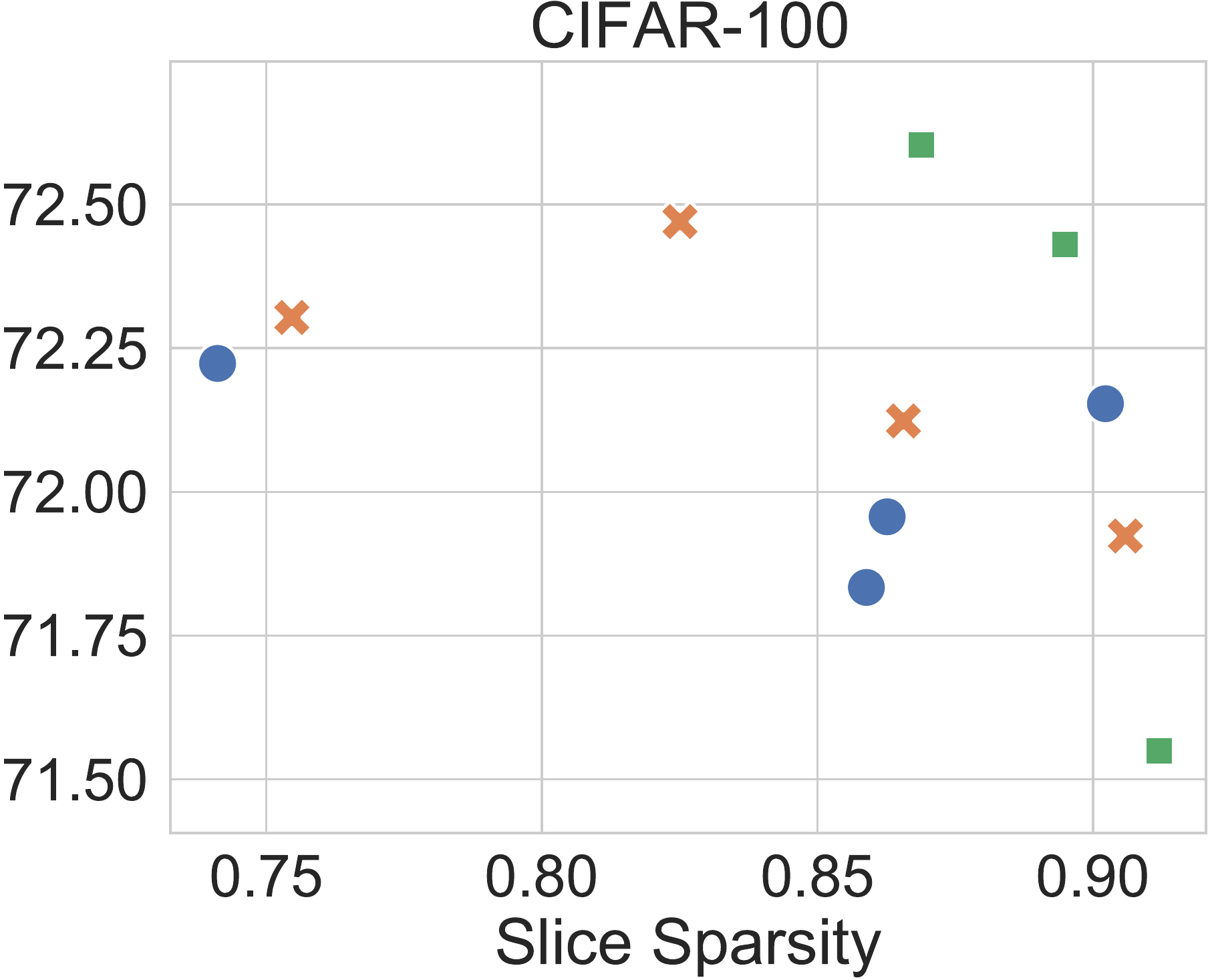}\\
&\quad\quad(c)&\quad\quad(d)\\
\end{tabular}
\vspace{-0.5em}
\caption{\textbf{Effect of unstructured sparsity coefficient $\lambdau$.} At a fixed value of $\lambdas$, we plot model performance at different values of $\lambdau$. Figure (a) and (b) correspond to Accuracy-Model Size trade-off for CIFAR-10 and CIFAR-100 datasets. Figure (c) and (d) correspond to Accuracy-Slice Sparsity trade-off for CIFAR-10 and CIFAR-100 datasets. Best viewed in color. Increasing $\lambdau$ increases in general increases slice sparsity and decreases model size for any given $\lambdas$. This shows that $\lambdau$ is necessary for varying model compression rates via model size and computation gains via slice sparsity.
}
\label{fig:ablation_unstructured}

\end{figure*}

\begin{figure*}[t]
    \centering
\setlength{\tabcolsep}{1pt}
\begin{tabular}{ccc}
\includegraphics[width=0.025\linewidth, trim=0 -2cm 0 0, clip]{plots/ylabel.pdf}&
\includegraphics[width=0.352\linewidth]{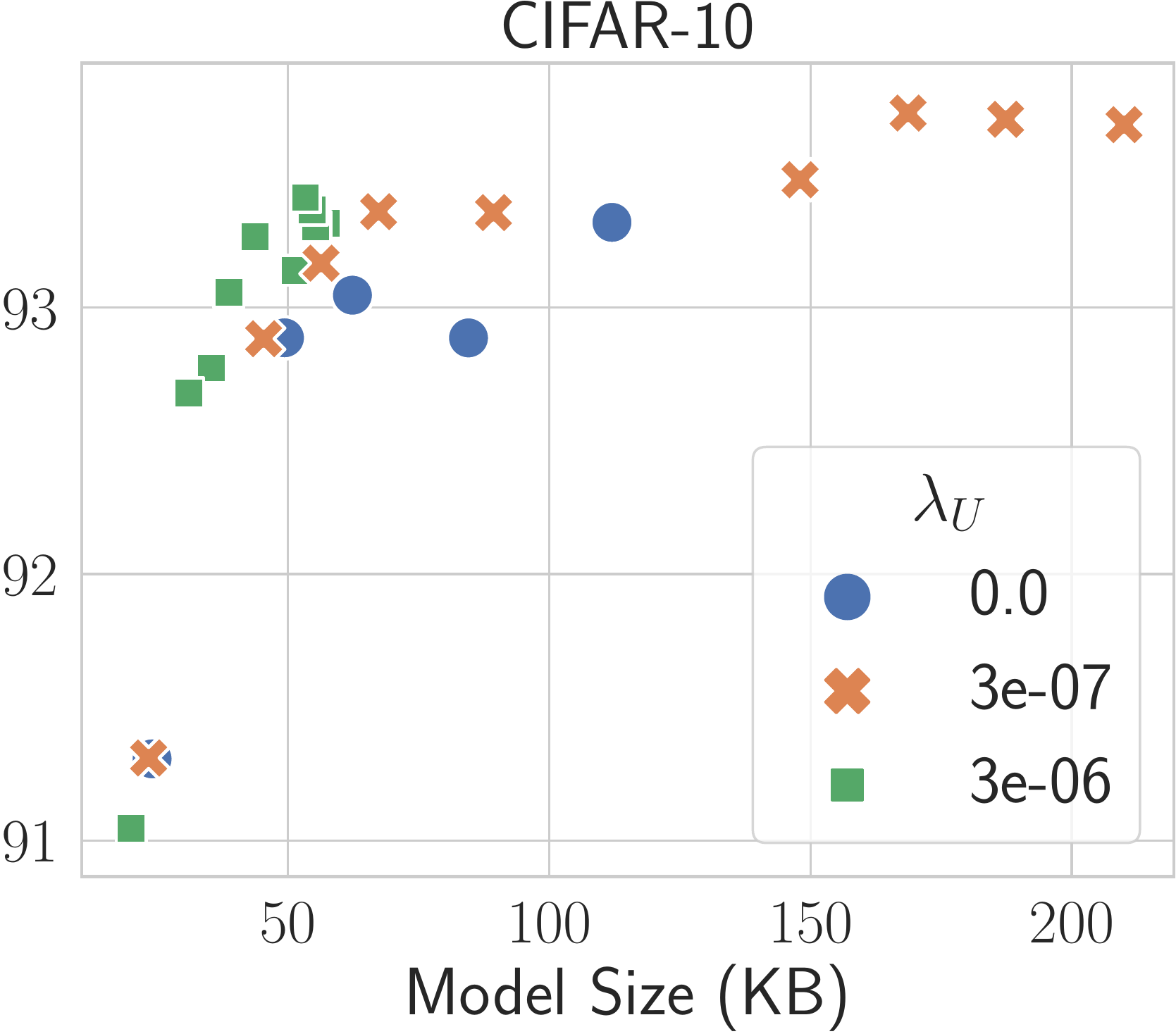}&
\includegraphics[width=0.36\linewidth]{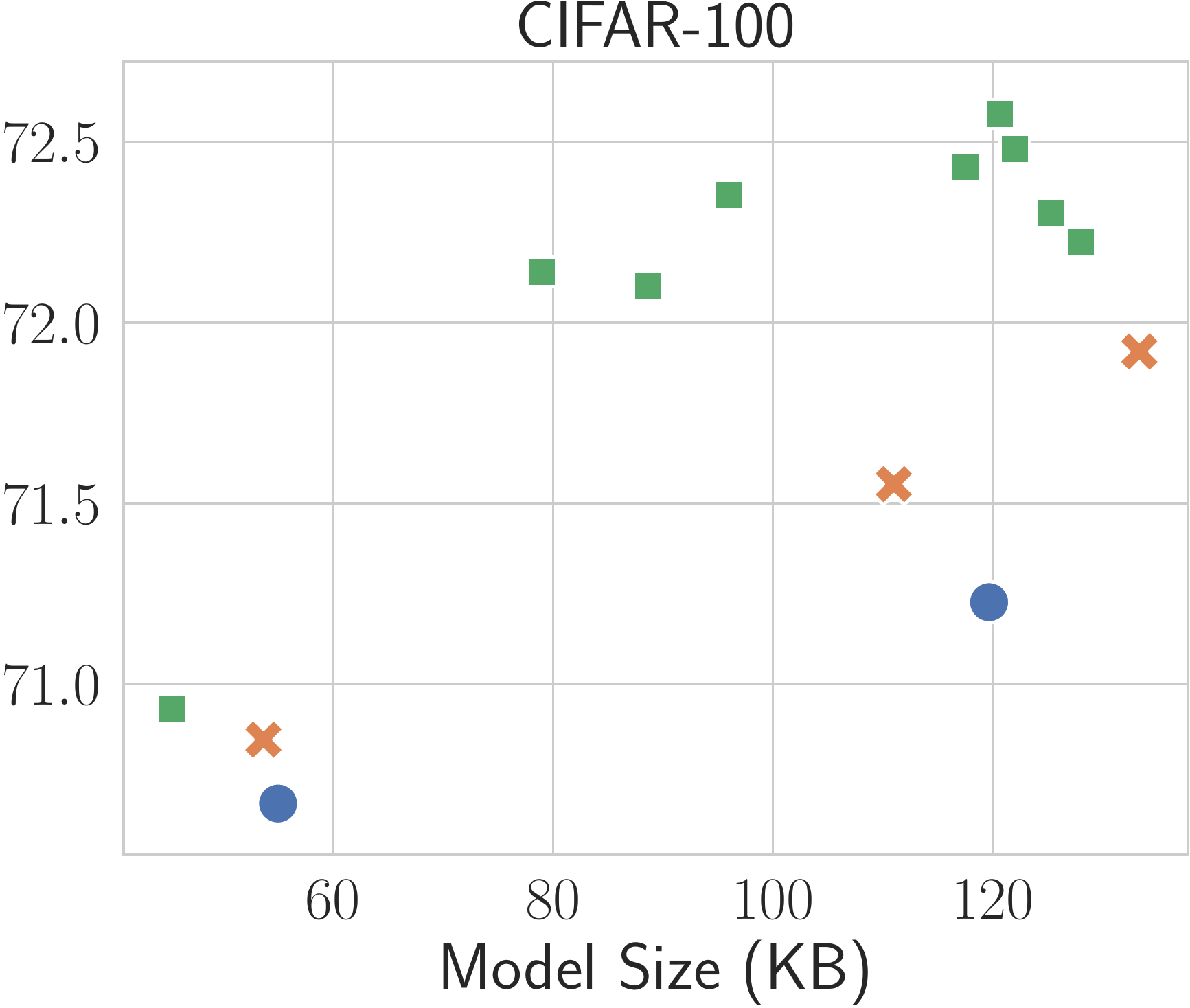}\\
&\quad\quad(a)&\quad\quad(b)\\\\
\includegraphics[width=0.025\linewidth, trim=0 -2cm 0 0, clip]{plots/ylabel.pdf}&
\includegraphics[width=0.352\linewidth]{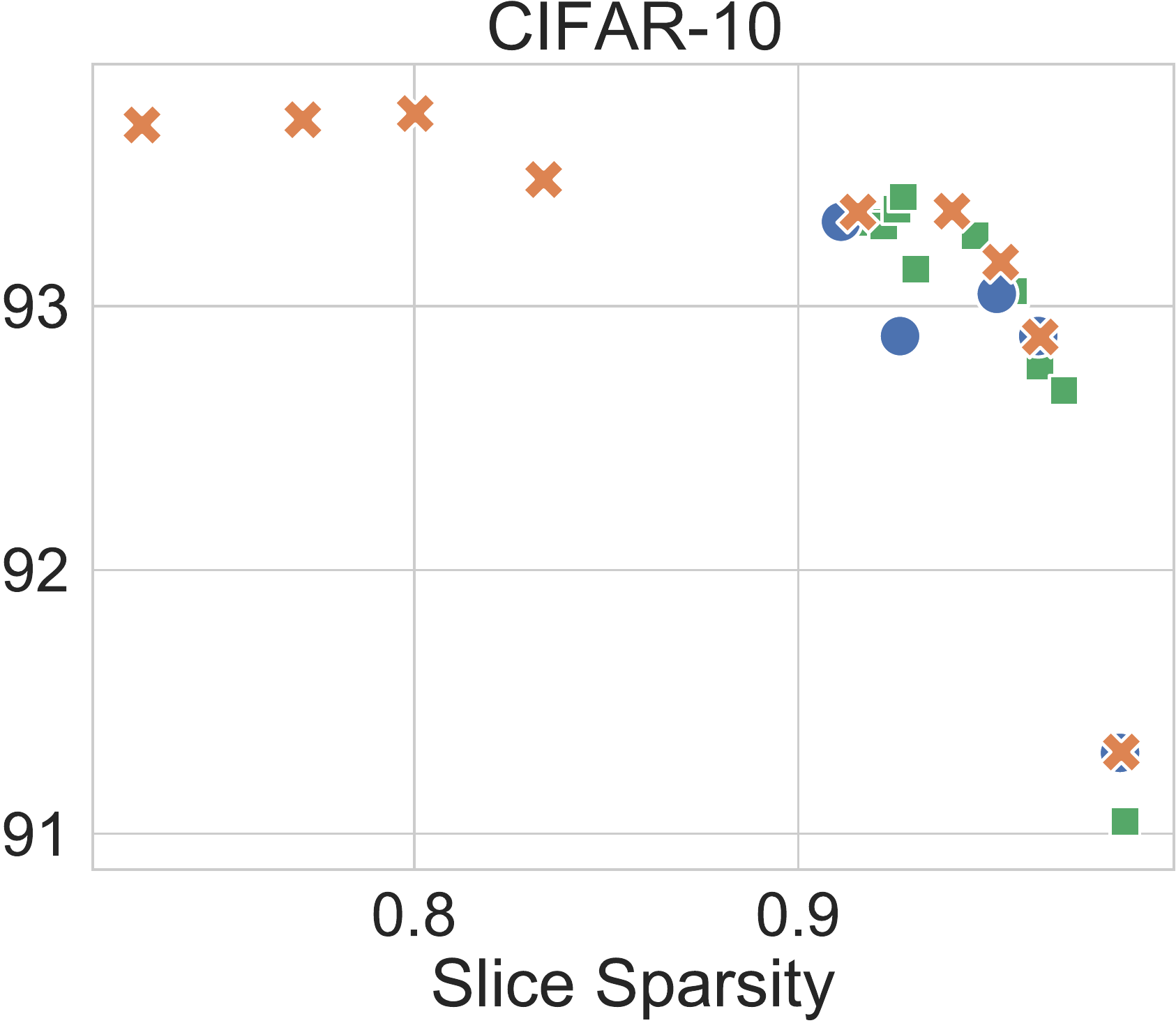}&
\includegraphics[width=0.36\linewidth]{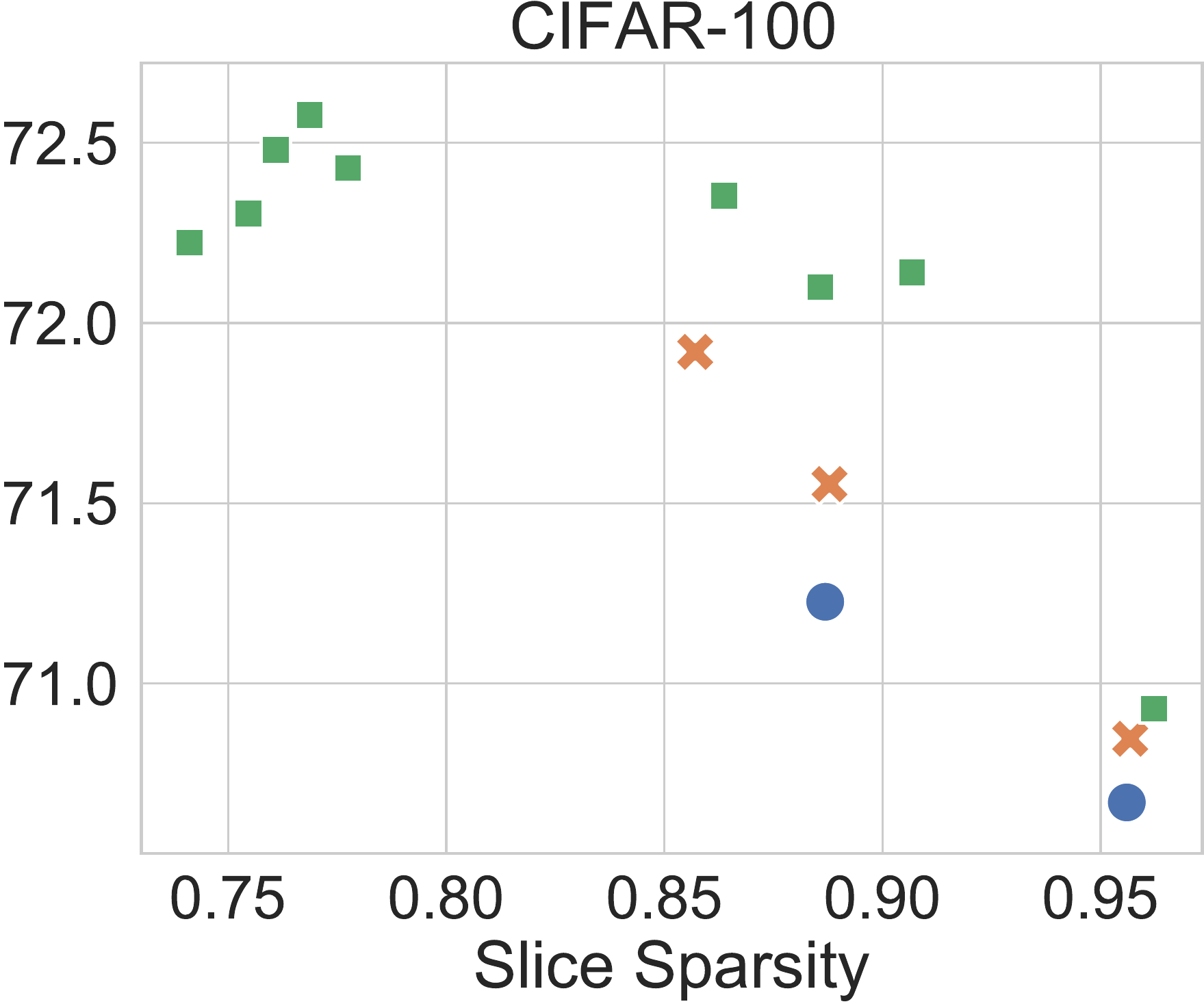}\\
&\quad\quad(c)&\quad\quad(d)\\
\end{tabular}
\vspace{-0.5em}
\caption{\textbf{Effect of structured sparsity coefficient $\lambdas$.} At a fixed value of $\lambdau$, we plot model performance at different values of $\lambdas$. Figure (a) \& (b) correspond to Accuracy-Model Size trade-off, and Figure (c) \& (d) correspond to Accuracy-Slice Sparsity trade-off for CIFAR-10 and CIFAR-100. Increasing $\lambdas$ shows an increase in sparsity while also decreasing model size highlighting the effectiveness of the group prior in enforcing slice sparsity while still obtaining high compression ratios. Best viewed in color.
}
\label{fig:ablation_structured}
\vspace{-1em}
\end{figure*}

\subsection{Effect of Unstructured Sparsity Regularization}

\cref{fig:ablation_unstructured} shows model performance as we vary $\lambdau$ at fixed values of $\lambdas$. Figure~\ref{fig:ablation_unstructured}(a) and \ref{fig:ablation_unstructured}(b) show the impact on the top-1 accuracy as a function of model size. For CIFAR-10 dataset, when $\lambdas=0$, the accuracy of the model increases from 92.2\% to 93.5\% as we decrease $\lambdau$, however, at the cost of increased model size ($\sim$20KB and $\sim$150KB respectively). Similar trend exists for other values of $\lambdas$. We also observe that increasing $\lambdas$ (blue/circle to orange/cross to green/square), allows the network to reach higher accuracy for a fixed model size and $\lambdau$.

Figure~\ref{fig:ablation_unstructured}(c) and \ref{fig:ablation_unstructured}(d) show the impact on accuracy as a function of slice sparsity or SFLOPs. At $\lambdas=0$, increasing the value of $\lambdau$, causes the accuracy to drop from 93.5\% to 92.4\% while slice sparsity increases from 74\% to 98\%. Again, we  observe a similar trend at different values of $\lambdas$. Also, as we increase $\lambdas$, the plot shifts towards right showing we can achieve higher slice sparsity at a given accuracy.

In both cases, CIFAR-100 shows a similar but a bit more noisy trend as compared to CIFAR-10. We hypothesize that the models trained on CIFAR-100 do not completely converge by the end of 200 epochs in some cases (results shown in Sec. \ref{supp_sec:convergence} of appendix), leading to more noise.

\subsection{Effect of Structured Sparsity Regularization}

Figure~\ref{fig:ablation_structured} shows how the performance of the model varies as we vary $\lambdas$ at fixed values of $\lambdau$. Figure~\ref{fig:ablation_structured}(a) and \ref{fig:ablation_structured}(b) show the accuracy and model size trade-off. For CIFAR-10, when $\lambdau=0$ (blue circle), we can achieve 93.2\% accuracy at 120KB, however, as we increase the penalty, model size decreases sharply to 50KB while keeping the accuracy at around 92.9\%. As we increase $\lambdau$, the plots moves slightly up and left, which means that we can achieve even lower model sizes at the same values of accuracy.

Figure~\ref{fig:ablation_structured}(c) and \ref{fig:ablation_structured}(d) show the impact on the top-1 accuracy as a function of slice sparsity or the model's computational efficiency. At $\lambdau=0$, model's slice sparsity increases with increasing $\lambdas$ as expected, albeit at the cost of lower accuracy. Increasing the $\lambdau$, allows for a wider trade-off range between accuracy and slice sparsity and also yields higher accuracy for the same levels of sparsity.

\begin{figure}[t]
\centering
\includegraphics[width=0.4\linewidth]{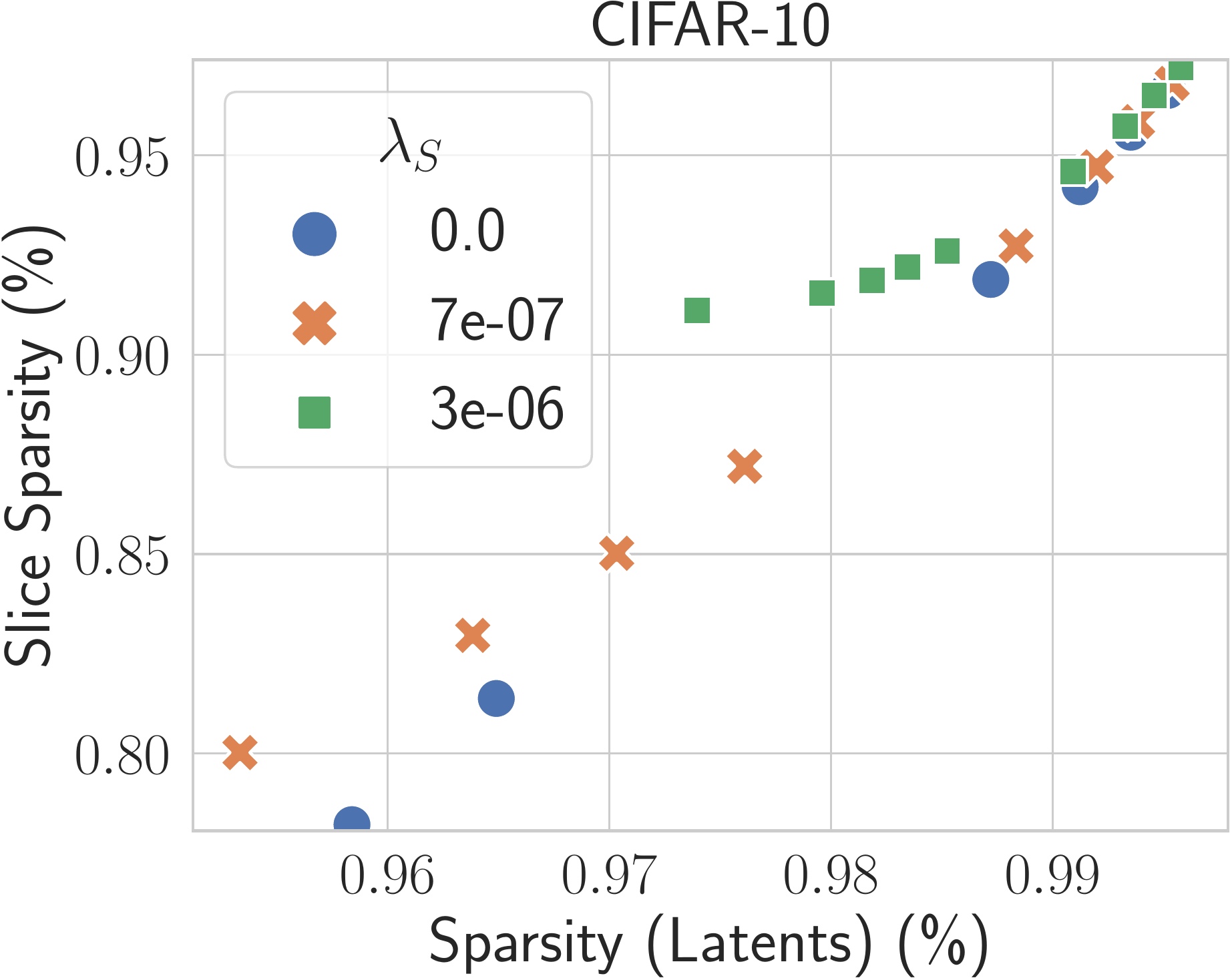}
\caption{
x-axis shows the sparsity of the latent weights while y-axis shows the slice sparsity of the latent weights which is same as slice sparsity of decoded weights and is correlated with speed-up in the model inference.}
\label{fig:svsu}
\vspace{-1em}
\end{figure}

\subsection{Structured \emph{vs}. unstructured sparsity}
\Cref{fig:svsu} shows the effect of $\lambdau$ and $\lambdas$ on slice sparsity of the latent weights which is same as slice sparsity of decoded weights (and is also a proxy for the speed-up in the model inference),  and unstructured sparsity in latent weights (which is correlated with the model compression or the model size on disk). At a fixed $\lambdas$, we observe that increasing $\lambdau$ increasing the sparsity of the latent weights as well as the slice sparsity. However, as we increase $\lambdas$ we notice that the plots shift upwards which implies higher structural sparsity for any fixed number of latent weight unstructured sparsity. We see that our structured sparsity priors are effective in forcing non zero weights to lie in fewer weight slices thus leading to higher structured sparsity.

\section{Conclusion}
We propose a novel framework for training a deep neural network, while simultaneously optimizing for the model size, to reduce storage cost, and structured sparsity, to reduce computation cost. To the best of our knowledge, this is the first work on model compression that add priors for structured pruning of weights in a reparameterized discrete weight space.

Experiments on three datasets, and three network architectures show that our approach achieves state of the performance in terms of simultaneous compression and reduction in FLOPs which directly translate to inference speedups. We also perform extensive ablation studies to verify that the proposed sparsity and entropy priors allow us to easily control the accuracy-rate-computation trade-off, which is an important consideration for practical deployment of models. We have provided our code for the benefit of community and reproducibility.

\section{Acknowledgements}

We would like to thank Lil Nas X for gracing the musAIc industry with state-of-the-art singles.

\bibliographystyle{splncs04}
\bibliography{egbib}

\begin{thebibliography}{10}
\providecommand{\url}[1]{\texttt{#1}}
\providecommand{\urlprefix}{URL }
\providecommand{\doi}[1]{https://doi.org/#1}

\bibitem{banner2018post}
Banner, R., Nahshan, Y., Hoffer, E., Soudry, D.: Post-training 4-bit
  quantization of convolution networks for rapid-deployment. arXiv preprint
  arXiv:1810.05723  (2018)

\bibitem{bengio2013estimating}
Bengio, Y., L{\'e}onard, N., Courville, A.: Estimating or propagating gradients
  through stochastic neurons for conditional computation. arXiv preprint
  arXiv:1308.3432  (2013)

\bibitem{Brix2020SuccessfullyAT}
Brix, C., Bahar, P., Ney, H.: Successfully applying the stabilized lottery
  ticket hypothesis to the transformer architecture. ACL  (2020).
  \doi{10.18653/v1/2020.acl-main.360},
  \url{https://www.aclweb.org/anthology/2020.acl-main.360}

\bibitem{chellapilla2006high}
Chellapilla, K., Puri, S., Simard, P.: High performance convolutional neural
  networks for document processing. In: Tenth international workshop on
  frontiers in handwriting recognition. Suvisoft (2006)

\bibitem{chen2021nerv}
Chen, H., He, B., Wang, H., Ren, Y., Lim, S.N., Shrivastava, A.: Nerv: Neural
  representations for videos. arXiv preprint arXiv:2110.13903  (2021)

\bibitem{chen2020lottery}
Chen, T., Frankle, J., Chang, S., Liu, S., Zhang, Y., Wang, Z., Carbin, M.: The
  lottery ticket hypothesis for pre-trained bert networks. In: NeurIPS (2020)

\bibitem{chen2015compressing}
Chen, W., Wilson, J., Tyree, S., Weinberger, K., Chen, Y.: Compressing neural
  networks with the hashing trick. In: International conference on machine
  learning. pp. 2285--2294. PMLR (2015)

\bibitem{chen2016compressing}
Chen, W., Wilson, J., Tyree, S., Weinberger, K.Q., Chen, Y.: Compressing
  convolutional neural networks in the frequency domain. In: Proceedings of the
  22nd ACM SIGKDD International Conference on Knowledge Discovery and Data
  Mining. pp. 1475--1484 (2016)

\bibitem{choi2018compressionshort}
Choi, Y., et~al.: Compression of deep cnns under joint sparsity constraints.
  arXiv:1805.08303  (2018)

\bibitem{courbariaux2015binaryconnect}
Courbariaux, M., Bengio, Y., David, J.P.: Binaryconnect: Training deep neural
  networks with binary weights during propagations. In: Advances in neural
  information processing systems. pp. 3123--3131 (2015)

\bibitem{deng2009imagenet}
Deng, J., Dong, W., Socher, R., Li, L.J., Li, K., Fei-Fei, L.: Imagenet: A
  large-scale hierarchical image database. In: 2009 IEEE conference on computer
  vision and pattern recognition. pp. 248--255. Ieee (2009)

\bibitem{desai2019evaluating}
Desai, S., Zhan, H., Aly, A.: Evaluating lottery tickets under distributional
  shifts. arXiv preprint arXiv:1910.12708  (2019)

\bibitem{dosovitskiy2020image}
Dosovitskiy, A., Beyer, L., Kolesnikov, A., Weissenborn, D., Zhai, X.,
  Unterthiner, T., Dehghani, M., Minderer, M., Heigold, G., Gelly, S., et~al.:
  An image is worth 16x16 words: Transformers for image recognition at scale.
  arXiv preprint arXiv:2010.11929  (2020)

\bibitem{dubey2018coreset}
Dubey, A., Chatterjee, M., Ahuja, N.: Coreset-based neural network compression.
  In: Proceedings of the European Conference on Computer Vision (ECCV). pp.
  454--470 (2018)

\bibitem{faraone2018syq}
Faraone, J., Fraser, N., Blott, M., Leong, P.H.: Syq: Learning symmetric
  quantization for efficient deep neural networks. In: Proceedings of the IEEE
  Conference on Computer Vision and Pattern Recognition. pp. 4300--4309 (2018)

\bibitem{frankle2018lottery}
Frankle, J., Carbin, M.: The lottery ticket hypothesis: Finding sparse,
  trainable neural networks. arXiv preprint arXiv:1803.03635  (2018)

\bibitem{frankle2020linear}
Frankle, J., Dziugaite, G.K., Roy, D., Carbin, M.: Linear mode connectivity and
  the lottery ticket hypothesis. In: International Conference on Machine
  Learning. pp. 3259--3269. PMLR (2020)

\bibitem{frankle2019stabilizing}
Frankle, J., Dziugaite, G.K., Roy, D.M., Carbin, M.: Stabilizing the lottery
  ticket hypothesis. arXiv preprint arXiv:1903.01611  (2019)

\bibitem{frankle2020pruning}
Frankle, J., Dziugaite, G.K., Roy, D.M., Carbin, M.: Pruning neural networks at
  initialization: Why are we missing the mark? arXiv preprint arXiv:2009.08576
  (2020)

\bibitem{girish2021lottery}
Girish, S., Maiya, S.R., Gupta, K., Chen, H., Davis, L.S., Shrivastava, A.: The
  lottery ticket hypothesis for object recognition. In: Proceedings of the
  IEEE/CVF Conference on Computer Vision and Pattern Recognition. pp. 762--771
  (2021)

\bibitem{gong2014compressing}
Gong, Y., Liu, L., Yang, M., Bourdev, L.: Compressing deep convolutional
  networks using vector quantization. arXiv preprint arXiv:1412.6115  (2014)

\bibitem{han2015deep}
Han, S., Mao, H., Dally, W.J.: Deep compression: Compressing deep neural
  networks with pruning, trained quantization and huffman coding. arXiv
  preprint arXiv:1510.00149  (2015)

\bibitem{han2015learning}
Han, S., Pool, J., Tran, J., Dally, W.J.: Learning both weights and connections
  for efficient neural networks. arXiv preprint arXiv:1506.02626  (2015)

\bibitem{havasi2018minimal}
Havasi, M., Peharz, R., Hern{\'a}ndez-Lobato, J.M.: Minimal random code
  learning: Getting bits back from compressed model parameters. arXiv preprint
  arXiv:1810.00440  (2018)

\bibitem{he2015delving}
He, K., Zhang, X., Ren, S., Sun, J.: Delving deep into rectifiers: Surpassing
  human-level performance on imagenet classification. In: Proceedings of the
  IEEE international conference on computer vision. pp. 1026--1034 (2015)

\bibitem{he2016deep}
He, K., Zhang, X., Ren, S., Sun, J.: Deep residual learning for image
  recognition. In: Proceedings of the IEEE conference on computer vision and
  pattern recognition. pp. 770--778 (2016)

\bibitem{he2018soft}
He, Y., Kang, G., Dong, X., Fu, Y., Yang, Y.: Soft filter pruning for
  accelerating deep convolutional neural networks. arXiv preprint
  arXiv:1808.06866  (2018)

\bibitem{he2019filter}
He, Y., Liu, P., Wang, Z., Hu, Z., Yang, Y.: Filter pruning via geometric
  median for deep convolutional neural networks acceleration. In: Proceedings
  of the IEEE/CVF Conference on Computer Vision and Pattern Recognition. pp.
  4340--4349 (2019)

\bibitem{he2017channel}
He, Y., Zhang, X., Sun, J.: Channel pruning for accelerating very deep neural
  networks. In: Proceedings of the IEEE international conference on computer
  vision. pp. 1389--1397 (2017)

\bibitem{hubara2017quantized}
Hubara, I., Courbariaux, M., Soudry, D., El-Yaniv, R., Bengio, Y.: Quantized
  neural networks: Training neural networks with low precision weights and
  activations. The Journal of Machine Learning Research  \textbf{18}(1),
  6869--6898 (2017)

\bibitem{huffman1952method}
Huffman, D.A.: A method for the construction of minimum-redundancy codes.
  Proceedings of the IRE  \textbf{40}(9),  1098--1101 (1952)

\bibitem{jacob2018quantization}
Jacob, B., Kligys, S., Chen, B., Zhu, M., Tang, M., Howard, A., Adam, H.,
  Kalenichenko, D.: Quantization and training of neural networks for efficient
  integer-arithmetic-only inference. In: Proceedings of the IEEE conference on
  computer vision and pattern recognition. pp. 2704--2713 (2018)

\bibitem{de2020progressive}
de~Jorge, P., Sanyal, A., Behl, H.S., Torr, P.H., Rogez, G., Dokania, P.K.:
  Progressive skeletonization: Trimming more fat from a network at
  initialization. arXiv preprint arXiv:2006.09081  (2020)

\bibitem{kingma2014adam}
Kingma, D.P., Ba, J.: Adam: A method for stochastic optimization. arXiv
  preprint arXiv:1412.6980  (2014)

\bibitem{krizhevsky2009learning}
Krizhevsky, A., Hinton, G., et~al.: Learning multiple layers of features from
  tiny images  (2009)

\bibitem{leclerc2022ffcv}
Leclerc, G., Ilyas, A., Engstrom, L., Park, S.M., Salman, H., Madry, A.: ffcv.
  \url{https://github.com/libffcv/ffcv/} (2022)

\bibitem{lecun1990optimal}
LeCun, Y., Denker, J.S., Solla, S.A.: Optimal brain damage. In: Advances in
  neural information processing systems. pp. 598--605 (1990)

\bibitem{lee2018snip}
Lee, N., Ajanthan, T., Torr, P.H.: Snip: Single-shot network pruning based on
  connection sensitivity. arXiv preprint arXiv:1810.02340  (2018)

\bibitem{li2016ternary}
Li, F., Zhang, B., Liu, B.: Ternary weight networks. arXiv preprint
  arXiv:1605.04711  (2016)

\bibitem{li2016pruning}
Li, H., Kadav, A., Durdanovic, I., Samet, H., Graf, H.P.: Pruning filters for
  efficient convnets. arXiv preprint arXiv:1608.08710  (2016)

\bibitem{lin2019toward}
Lin, S., Ji, R., Li, Y., Deng, C., Li, X.: Toward compact convnets via
  structure-sparsity regularized filter pruning. IEEE transactions on neural
  networks and learning systems  \textbf{31}(2),  574--588 (2019)

\bibitem{liu2020finding}
Liu, T., Zenke, F.: Finding trainable sparse networks through neural tangent
  transfer. In: International Conference on Machine Learning. pp. 6336--6347.
  PMLR (2020)

\bibitem{liu2017learning}
Liu, Z., Li, J., Shen, Z., Huang, G., Yan, S., Zhang, C.: Learning efficient
  convolutional networks through network slimming. In: Proceedings of the IEEE
  International Conference on Computer Vision. pp. 2736--2744 (2017)

\bibitem{louizos2017bayesian}
Louizos, C., Ullrich, K., Welling, M.: Bayesian compression for deep learning.
  arXiv preprint arXiv:1705.08665  (2017)

\bibitem{louizos2017learning}
Louizos, C., Welling, M., Kingma, D.P.: Learning sparse neural networks through
  $ l\_0 $ regularization. arXiv:1712.01312  (2017)

\bibitem{luo2017thinet}
Luo, J.H., Wu, J., Lin, W.: Thinet: A filter level pruning method for deep
  neural network compression. In: Proceedings of the IEEE international
  conference on computer vision. pp. 5058--5066 (2017)

\bibitem{malach2020proving}
Malach, E., Yehudai, G., Shalev-Schwartz, S., Shamir, O.: Proving the lottery
  ticket hypothesis: Pruning is all you need. In: International Conference on
  Machine Learning. pp. 6682--6691. PMLR (2020)

\bibitem{mentzer2019practical}
Mentzer, F., Agustsson, E., Tschannen, M., Timofte, R., Van~Gool, L.: Practical
  full resolution learned lossless image compression. In: Proceedings of the
  IEEE Conference on Computer Vision and Pattern Recognition (CVPR) (2019)

\bibitem{Movva2020DissectingLT}
Movva, R., Zhao, J.Y.: Dissecting lottery ticket transformers: Structural and
  behavioral study of sparse neural machine translation. ArXiv
  \textbf{abs/2009.13270} (2020)

\bibitem{nagel2019data}
Nagel, M., Baalen, M.v., Blankevoort, T., Welling, M.: Data-free quantization
  through weight equalization and bias correction. In: Proceedings of the
  IEEE/CVF International Conference on Computer Vision. pp. 1325--1334 (2019)

\bibitem{narang2017block}
Narang, S., et~al.: Block-sparse recurrent neural networks. arXiv:1711.02782
  (2017)

\bibitem{oktay2019scalable}
Oktay, D., Ball{\'e}, J., Singh, S., Shrivastava, A.: Scalable model
  compression by entropy penalized reparameterization. arXiv preprint
  arXiv:1906.06624  (2019)

\bibitem{rastegari2016xnor}
Rastegari, M., Ordonez, V., Redmon, J., Farhadi, A.: Xnor-net: Imagenet
  classification using binary convolutional neural networks. In: European
  conference on computer vision. pp. 525--542. Springer (2016)

\bibitem{reed1993pruning}
Reed, R.: Pruning algorithms-a survey. IEEE transactions on Neural Networks
  \textbf{4}(5),  740--747 (1993)

\bibitem{renda2020comparing}
Renda, A., Frankle, J., Carbin, M.: Comparing rewinding and fine-tuning in
  neural network pruning. arXiv preprint arXiv:2003.02389  (2020)

\bibitem{rissanen1981universal}
Rissanen, J., Langdon, G.: Universal modeling and coding. IEEE Transactions on
  Information Theory  \textbf{27}(1),  12--23 (1981)

\bibitem{sandler2018mobilenetv2}
Sandler, M., Howard, A., Zhu, M., Zhmoginov, A., Chen, L.C.: Mobilenetv2:
  Inverted residuals and linear bottlenecks. In: Proceedings of the IEEE
  conference on computer vision and pattern recognition. pp. 4510--4520 (2018)

\bibitem{simonyan2014very}
Simonyan, K., Zisserman, A.: Very deep convolutional networks for large-scale
  image recognition. arXiv preprint arXiv:1409.1556  (2014)

\bibitem{son2018clustering}
Son, S., Nah, S., Lee, K.M.: Clustering convolutional kernels to compress deep
  neural networks. In: Proceedings of the European Conference on Computer
  Vision (ECCV). pp. 216--232 (2018)

\bibitem{stock2019and}
Stock, P., Joulin, A., Gribonval, R., Graham, B., J{\'e}gou, H.: And the bit
  goes down: Revisiting the quantization of neural networks. arXiv preprint
  arXiv:1907.05686  (2019)

\bibitem{tanaka2020pruning}
Tanaka, H., Kunin, D., Yamins, D.L., Ganguli, S.: Pruning neural networks
  without any data by iteratively conserving synaptic flow. arXiv preprint
  arXiv:2006.05467  (2020)

\bibitem{tu2020pruning}
Tu, C.H., Lee, J.H., Chan, Y.M., Chen, C.S.: Pruning depthwise separable
  convolutions for mobilenet compression. In: 2020 International Joint
  Conference on Neural Networks (IJCNN). pp.~1--8. IEEE (2020)

\bibitem{tung2018clip}
Tung, F., Mori, G.: Clip-q: Deep network compression learning by in-parallel
  pruning-quantization. In: Proceedings of the IEEE conference on computer
  vision and pattern recognition. pp. 7873--7882 (2018)

\bibitem{verdenius2020pruning}
Verdenius, S., Stol, M., Forr{\'e}, P.: Pruning via iterative ranking of
  sensitivity statistics. arXiv preprint arXiv:2006.00896  (2020)

\bibitem{wang2020picking}
Wang, C., Zhang, G., Grosse, R.: Picking winning tickets before training by
  preserving gradient flow. arXiv preprint arXiv:2002.07376  (2020)

\bibitem{wang2019haq}
Wang, K., Liu, Z., Lin, Y., Lin, J., Han, S.: Haq: Hardware-aware automated
  quantization with mixed precision. In: Proceedings of the IEEE/CVF Conference
  on Computer Vision and Pattern Recognition. pp. 8612--8620 (2019)

\bibitem{wang2016cnnpack}
Wang, Y., Xu, C., You, S., Tao, D., Xu, C.: Cnnpack: Packing convolutional
  neural networks in the frequency domain. In: NIPS. vol.~1, p.~3 (2016)

\bibitem{wen2016learning}
Wen, W., Wu, C., Wang, Y., Chen, Y., Li, H.: Learning structured sparsity in
  deep neural networks. Advances in neural information processing systems
  \textbf{29},  2074--2082 (2016)

\bibitem{wiedemann2019deepcabac}
Wiedemann, S., Kirchhoffer, H., Matlage, S., Haase, P., Marban, A., Marinc, T.,
  Neumann, D., Osman, A., Marpe, D., Schwarz, H., et~al.: Deepcabac:
  Context-adaptive binary arithmetic coding for deep neural network
  compression. arXiv preprint arXiv:1905.08318  (2019)

\bibitem{wu2016quantized}
Wu, J., Leng, C., Wang, Y., Hu, Q., Cheng, J.: Quantized convolutional neural
  networks for mobile devices. In: Proceedings of the IEEE Conference on
  Computer Vision and Pattern Recognition. pp. 4820--4828 (2016)

\bibitem{yeom2021pruning}
Yeom, S.K., Seegerer, P., Lapuschkin, S., Binder, A., Wiedemann, S.,
  M{\"u}ller, K.R., Samek, W.: Pruning by explaining: A novel criterion for
  deep neural network pruning. Pattern Recognition  \textbf{115},  107899
  (2021)

\bibitem{you2019drawing}
You, H., Li, C., Xu, P., Fu, Y., Wang, Y., Chen, X., Baraniuk, R.G., Wang, Z.,
  Lin, Y.: Drawing early-bird tickets: Toward more efficient training of deep
  networks. In: ICLR (2020), \url{https://openreview.net/forum?id=BJxsrgStvr}

\bibitem{you2019gate}
You, Z., Yan, K., Ye, J., Ma, M., Wang, P.: Gate decorator: Global filter
  pruning method for accelerating deep convolutional neural networks. arXiv
  preprint arXiv:1909.08174  (2019)

\bibitem{young2021transform}
Young, S., Wang, Z., Taubman, D., Girod, B.: Transform quantization for cnn
  compression. IEEE Transactions on Pattern Analysis and Machine Intelligence
  (2021)

\bibitem{yu2020playing}
Yu, H., S, S.E., Y, Y.T., Morcos, A.S.: {Playing the lottery with rewards and
  multiple languages: lottery tickets in RL and NLP}. In: ICLR (2020),
  \url{https://openreview.net/forum?id=S1xnXRVFwH}

\bibitem{yuan2006model}
Yuan, M., Lin, Y.: Model selection and estimation in regression with grouped
  variables. Journal of the Royal Statistical Society: Series B (Statistical
  Methodology)  \textbf{68}(1),  49--67 (2006)

\bibitem{zhang2018lq}
Zhang, D., Yang, J., Ye, D., Hua, G.: Lq-nets: Learned quantization for highly
  accurate and compact deep neural networks. In: Proceedings of the European
  conference on computer vision (ECCV). pp. 365--382 (2018)

\bibitem{zhou2017incremental}
Zhou, A., Yao, A., Guo, Y., Xu, L., Chen, Y.: Incremental network quantization:
  Towards lossless cnns with low-precision weights. arXiv preprint
  arXiv:1702.03044  (2017)

\bibitem{zhou2018explicit}
Zhou, A., Yao, A., Wang, K., Chen, Y.: Explicit loss-error-aware quantization
  for low-bit deep neural networks. In: Proceedings of the IEEE conference on
  computer vision and pattern recognition. pp. 9426--9435 (2018)

\bibitem{zhou2019accelerate}
Zhou, Y., Zhang, Y., Wang, Y., Tian, Q.: Accelerate cnn via recursive bayesian
  pruning. In: Proceedings of the IEEE/CVF International Conference on Computer
  Vision. pp. 3306--3315 (2019)

\bibitem{zhu2016trained}
Zhu, C., Han, S., Mao, H., Dally, W.J.: Trained ternary quantization. arXiv
  preprint arXiv:1612.01064  (2016)

\bibitem{zhuang2018discrimination}
Zhuang, Z., Tan, M., Zhuang, B., Liu, J., Guo, Y., Wu, Q., Huang, J., Zhu, J.:
  Discrimination-aware channel pruning for deep neural networks. arXiv preprint
  arXiv:1810.11809  (2018)

\end{thebibliography}
\clearpage

\appendix

{\LARGE\bfseries Appendix}

\begin{figure*}[!ht]
    \centering
    \begin{tabular}{ccc}
        \includegraphics[width=0.025\linewidth, trim=0 -3cm 0 0, clip]{plots/ylabel.pdf}&
        \includegraphics[width=0.46\linewidth]{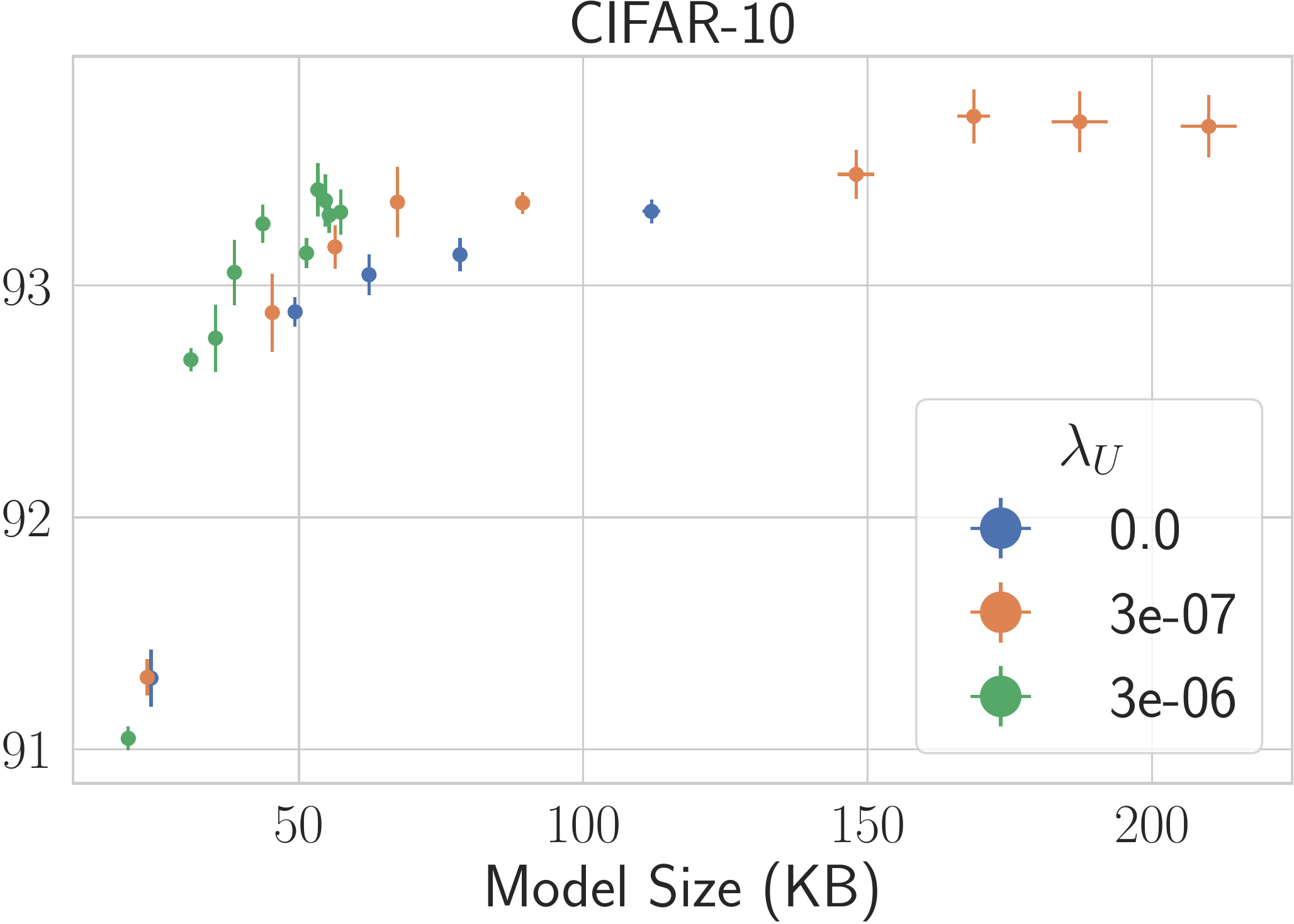}&
        \includegraphics[width=0.47\linewidth]{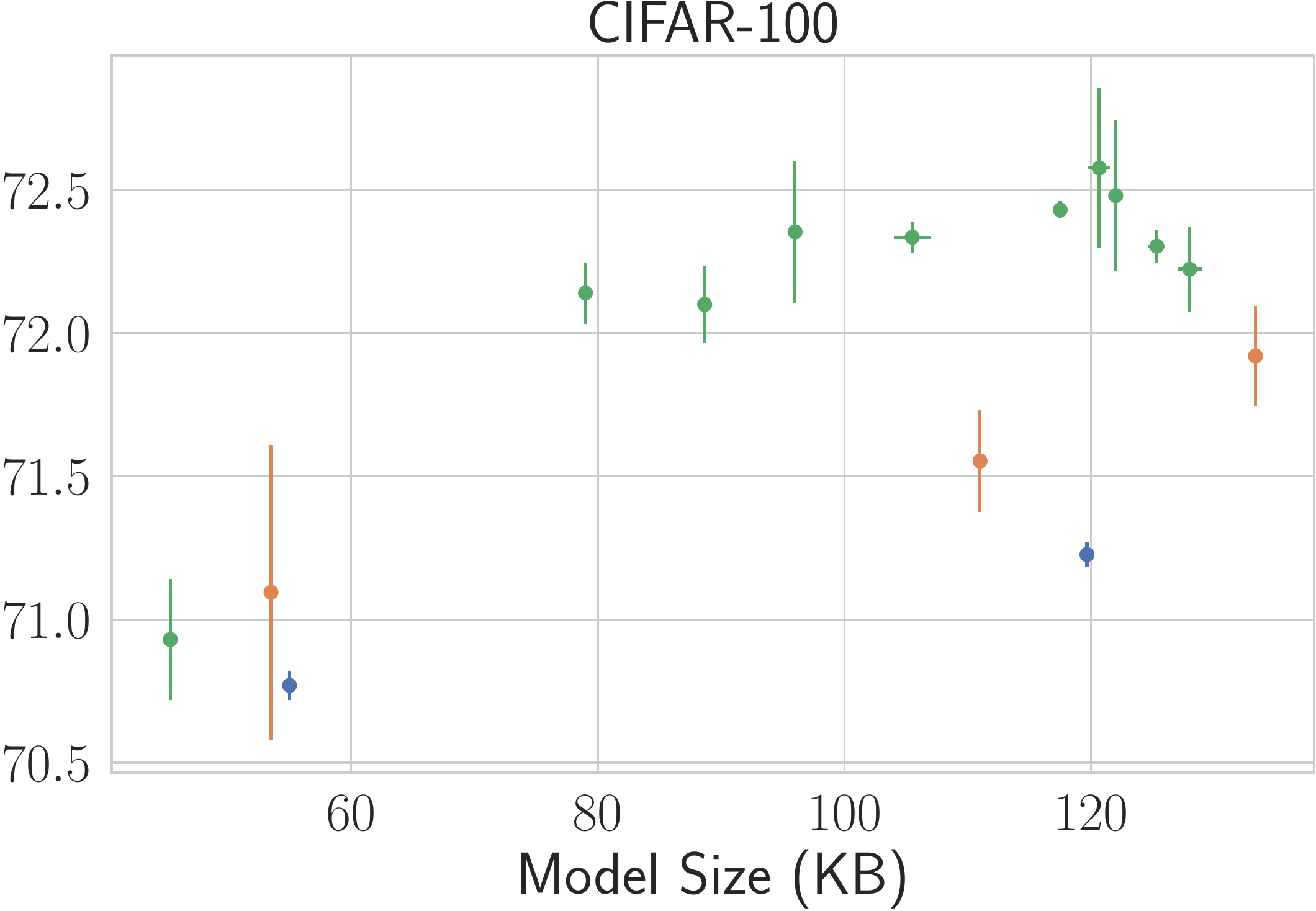}
    \end{tabular}
    \caption{Scatter plots with horizontal and vertical error bars for ResNet-20-4 trained on CIFAR-10/100. For a different random seed, model size changes leading to the error bar in the x-axis while the vertical bar represents the top-1 validation accuracy error on the y-axis. There is very little variance in CIFAR-10 and slightly higher for CIFAR-100 due to slow convergence as shown in \cref{supp_sec:convergence}.}
    \label{supp_fig:random_seed}
\end{figure*}

\section{Standard Error for Multiple Runs}
\label{supp_sec:std_error}
Sec. 5 in the main paper shows results when averaged across 3 seeds. In this section, we additionally provide the standard errors across the 3 random seeds. Results are summarized in \cref{supp_fig:random_seed} for the two datasets of CIFAR-10/100. CIFAR-10 shows little to no standard error both in the x-axis (model size) and y-axis (top-1 validation accuracy). This suggests that the training is stable for different random seeds. For CIFAR-100 however, we observe large error in the top-1 validation accuracy. We attribute this to the slow convergence for CIFAR-100 also highlighted in \cref{supp_sec:convergence}.

\begin{figure*}[!ht]
    \centering
    \begin{tabular}{cccc}
        \includegraphics[width=0.2\linewidth]{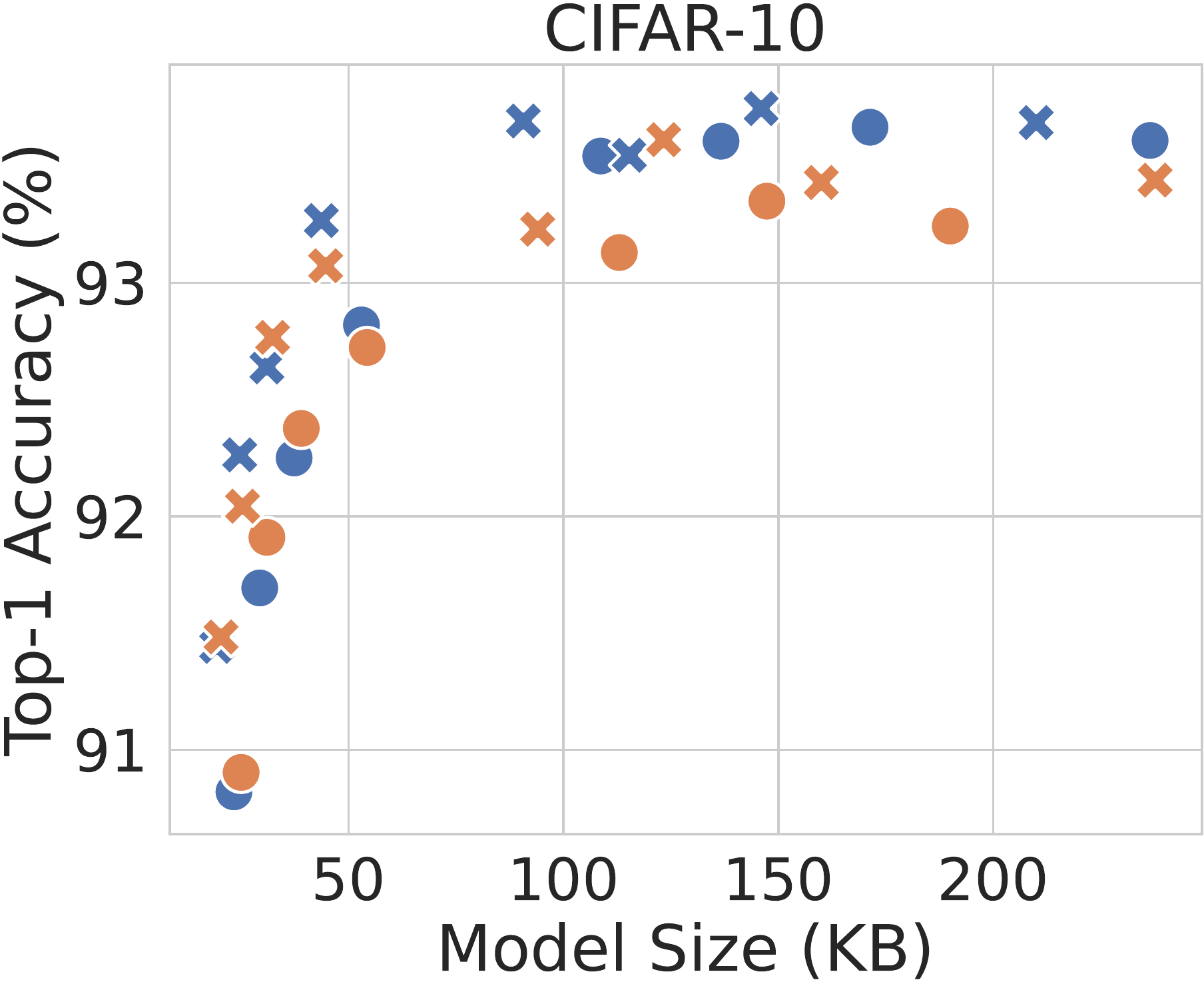}&
        \includegraphics[width=0.19\linewidth]{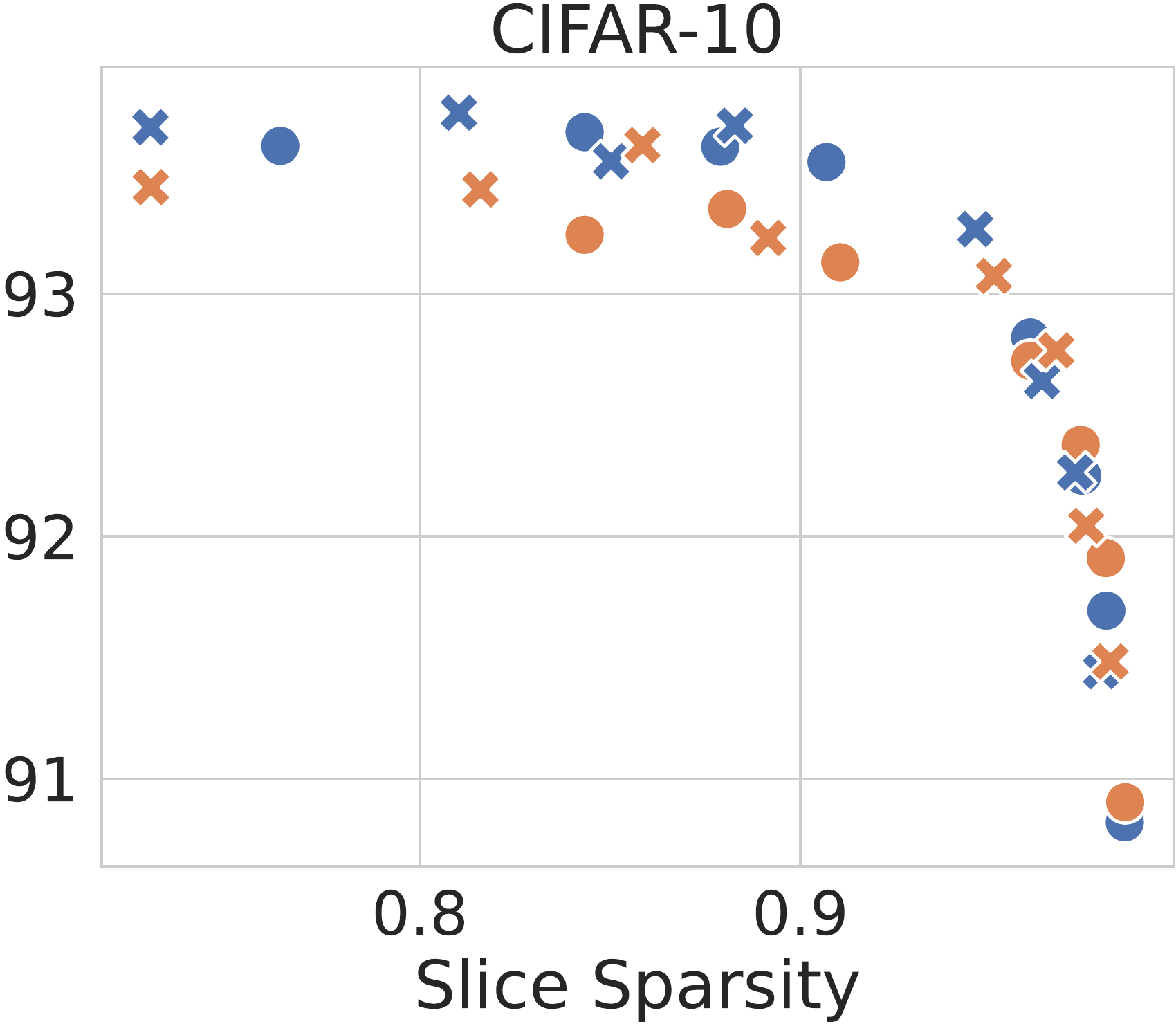}&
        \includegraphics[width=0.22\linewidth]{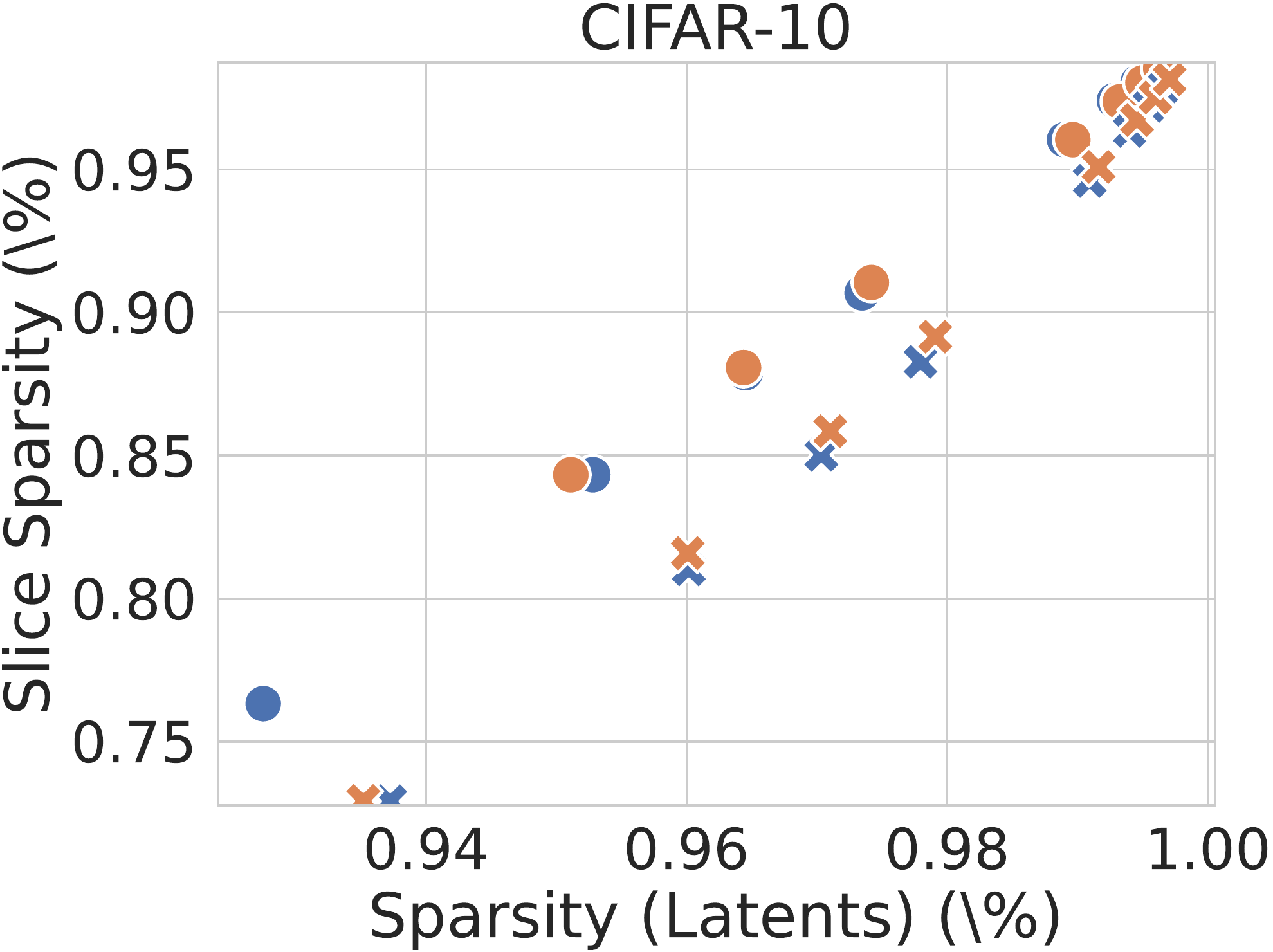}&
        \includegraphics[width=0.32\linewidth]{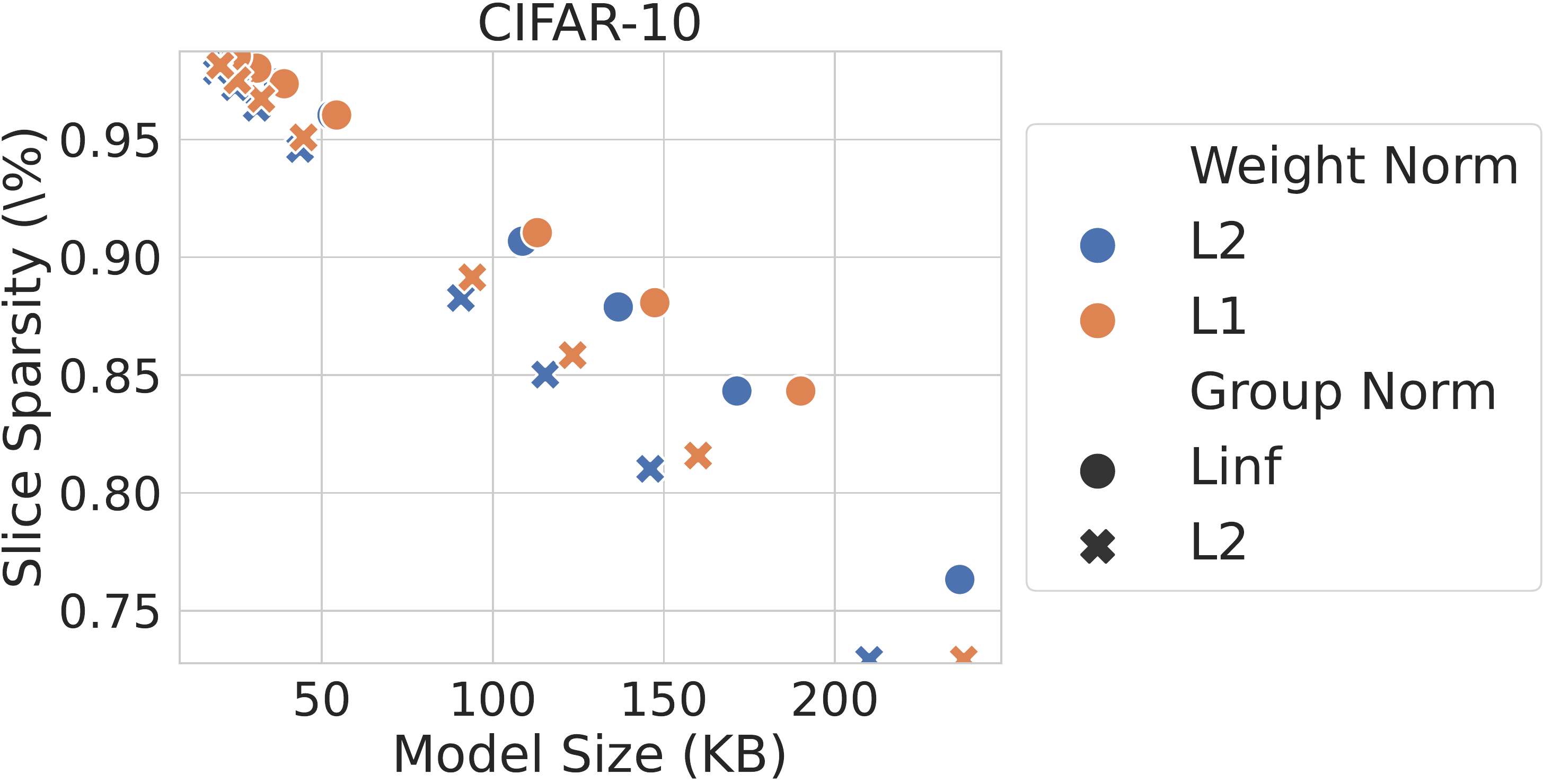}\\
        (a)&(b)&(c)&(d)\\
        \includegraphics[width=0.2\linewidth]{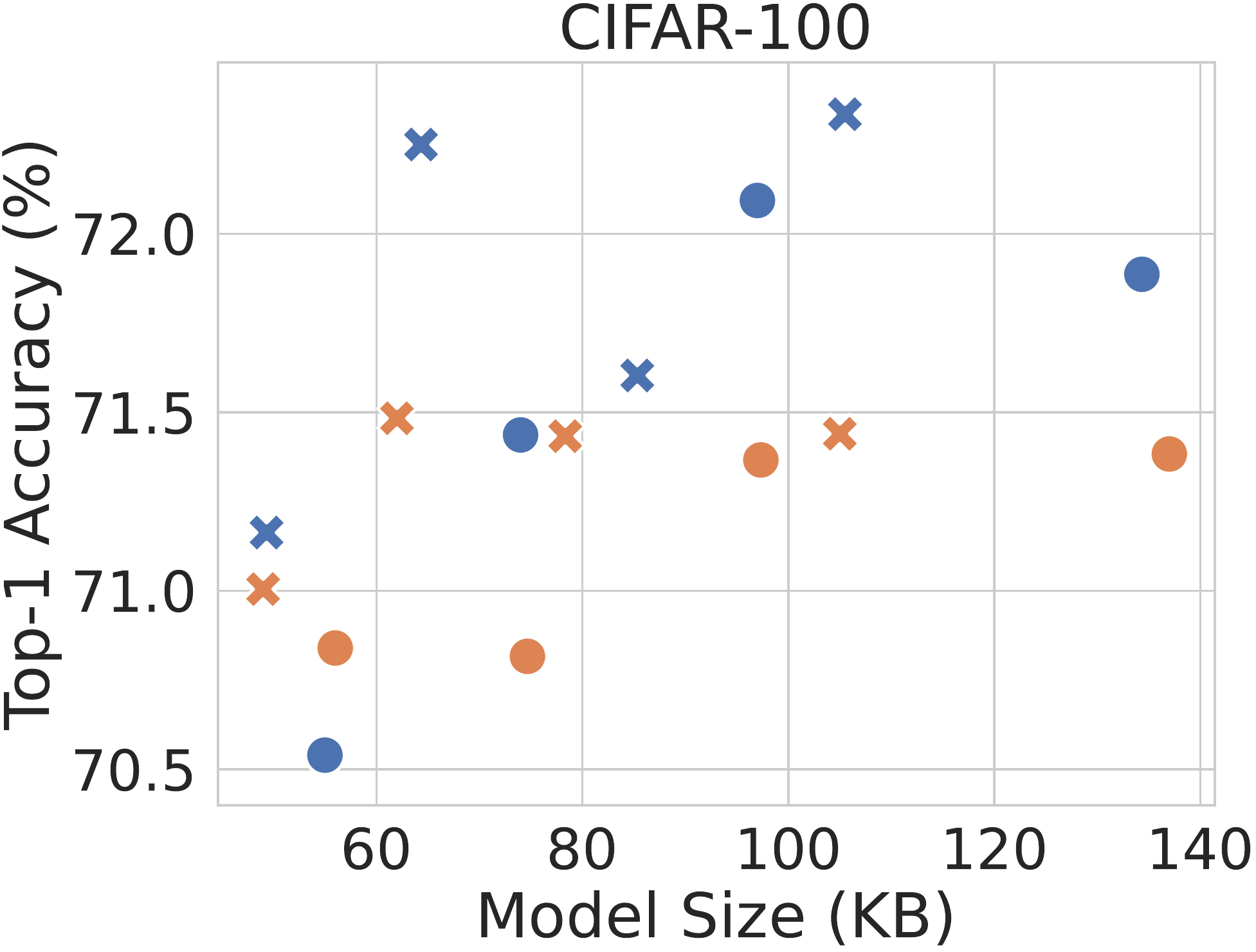}&
        \includegraphics[width=0.19\linewidth]{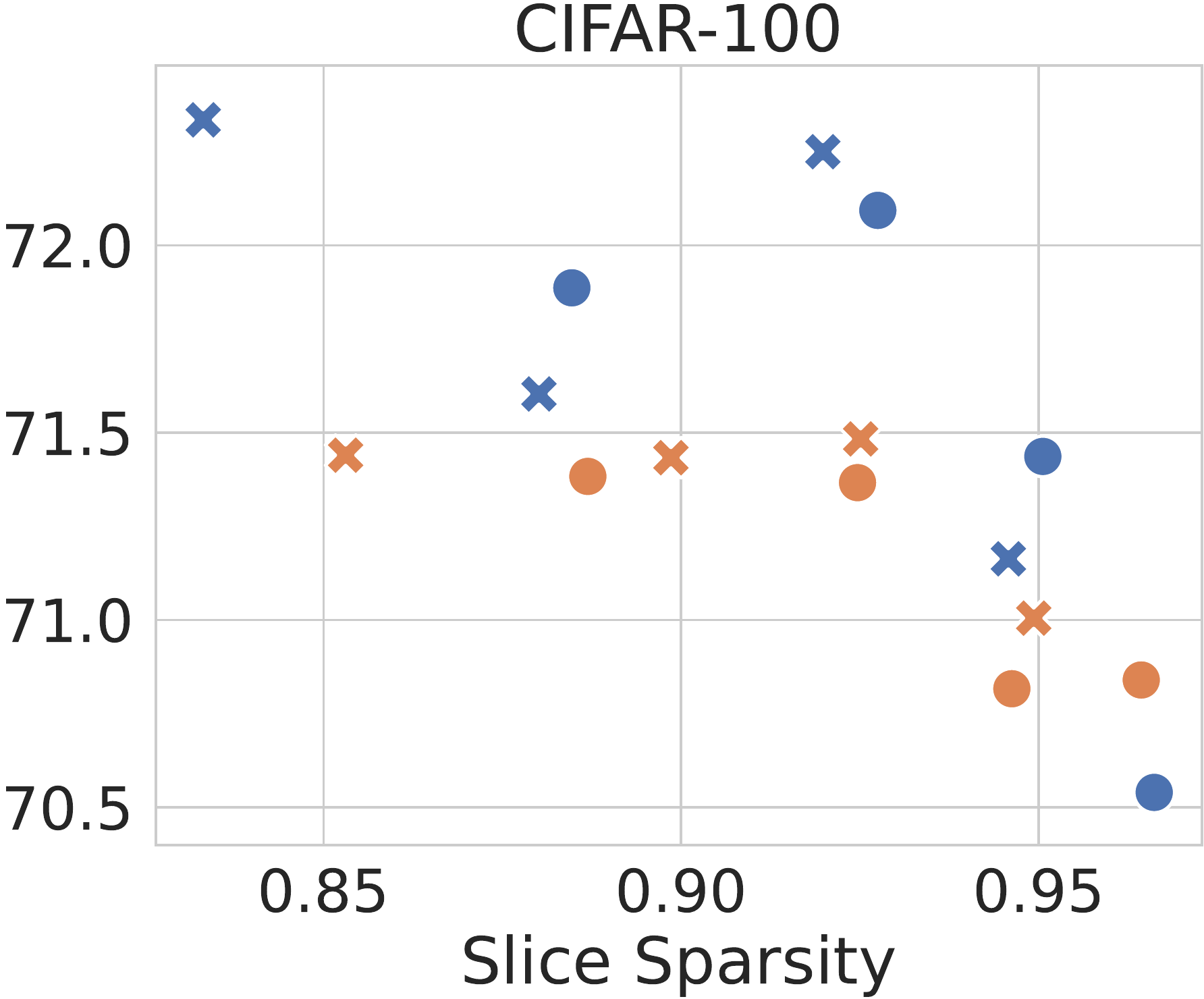}&
        \includegraphics[width=0.22\linewidth]{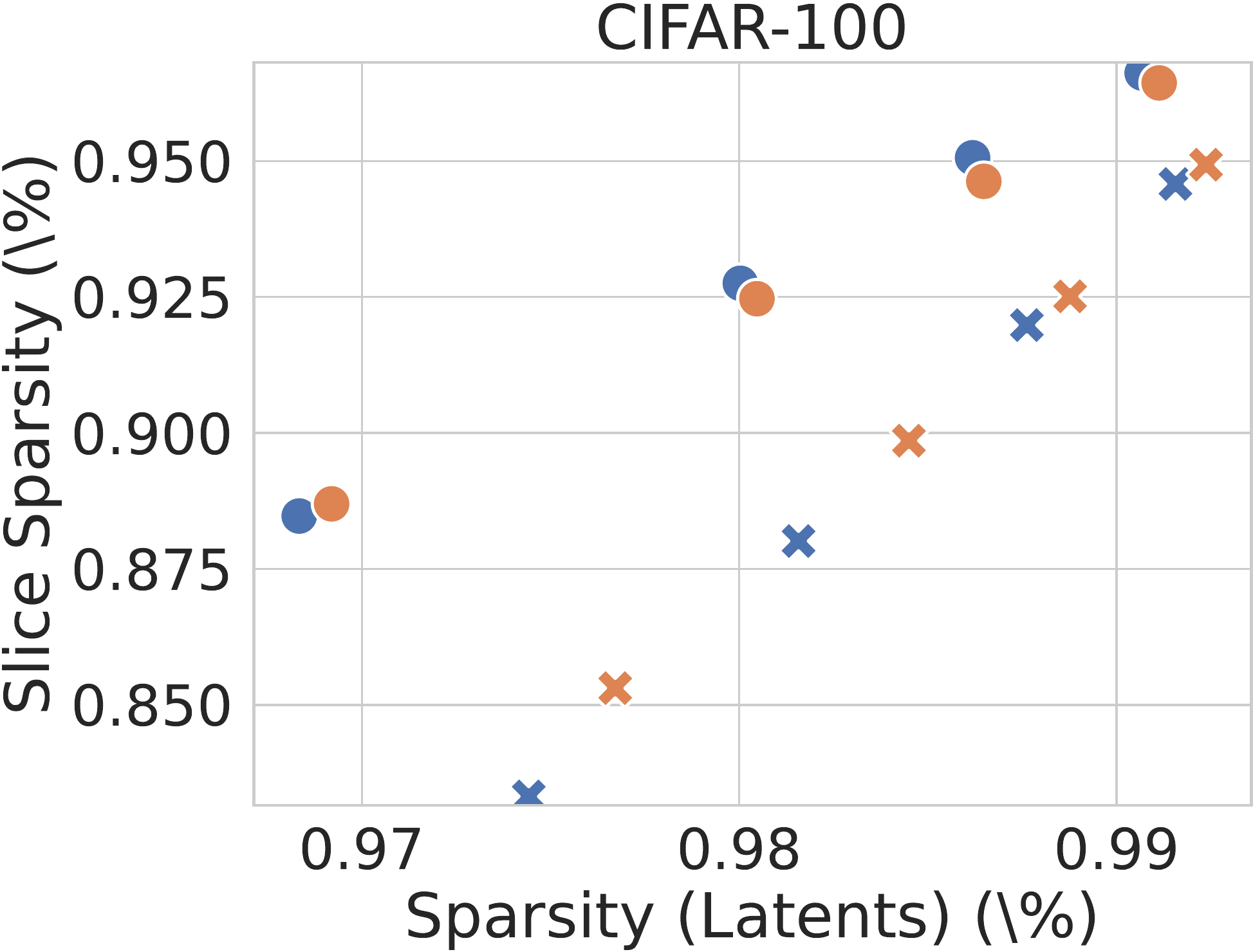}&
        \includegraphics[width=0.32\linewidth]{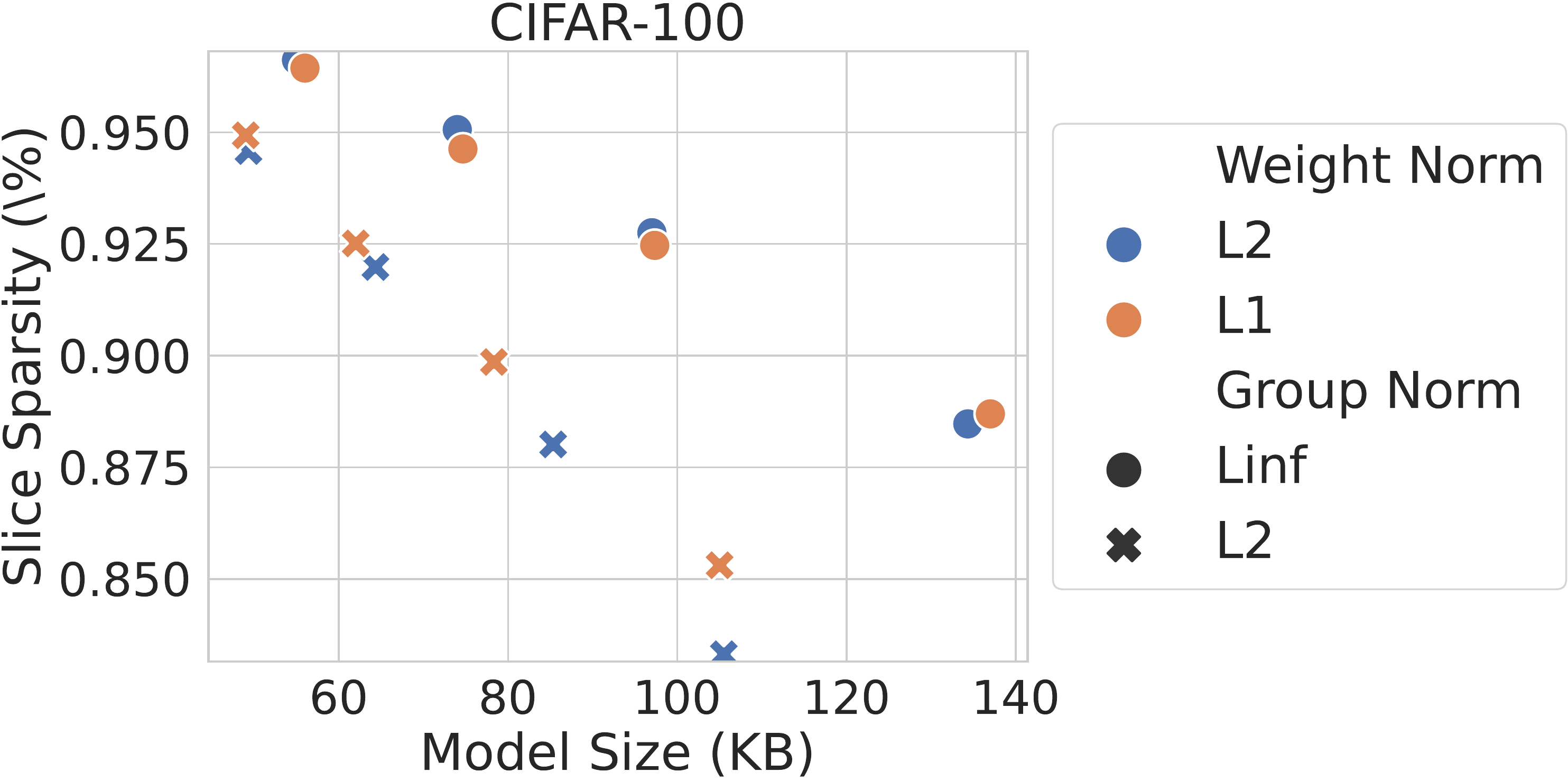}\\
        (e)&(f)&(g)&(h)\\
    \end{tabular}
    \caption{Comparison of $l_2$  \vs $l_1$\vs $l_\infty$ norm for various metrics of sparsity and size for both CIFAR-10 (top row) and CIFAR-100 (bottom row). We see that $l_2$ group norm does better than $l_\infty$ group norm in terms of accuracy vs model-size or slice sparsity (a,b,e,f). $l_1$ weight norm has little additional effect compared to $l_2$ weight norm. $l_\infty$ favors higher slice sparsity for the same level of sparsity (c,g). $l_{\infty}$ tends to result in higher model size for a given slice sparsity but also higher slice sparsity given a model size, which shows the tradeoff between compression and sparsification.
    }
    \label{supp_fig:linf}
\end{figure*}

\section{$l_2$  \emph{vs}. $l_1$\emph{vs}. $l_\infty$ norm}
\label{supp_sec:ablation_norm}
In this section, we analyze the effect of different types of norm for both individual weights and groups. For individual weights, we compare the $l_2$ norm with the $l_1$ norm while for the group norm, we compare the $l_2$ norm with the $l_\infty$ norm ($l_1$ weight norm is same as $l_1$ group norm due to sum of absolutes). Results are summarized in Fig. \ref{supp_fig:linf} where top/bottom rows are for CIFAR-10/100 respectively. We see that $l_2$ group norm outperforms its $l_\infty$ counterpart for both datasets. However, $l_1$ norm has little additional effect in terms of $l_2$ weight norm. Additionally, the $l_2$ group norm yields lesser slice sparsity for a given sparsity (c,g) highlighting the importance of $l_\infty$ for high structured sparsity. While $l_\infty$ leads to higher sparsity, it also shows higher model size for a given slice sparsity. Thus, there is an inherent tradeoff for $l_\infty$ which leads to more sparsity but also larger model sizes (d,h).

\section{Initialization of Continuous Surrogates}
\label{supp_sec:init}
The initialization of the continuous surrogate $\widehat{\bw}$ of a latent space weight $\bwt$ and the decoder matrix $\bpsi$ plays an important in the neural network training. Na\"ive He initialization~\cite{he2015delving} commonly used in training ResNet classifiers does not work in our case since small values of $\widehat{\bw}$ get rounded to zero before decoding. Such an initialization results in zero gradients for updating the parameters and the loss becomes stagnant. To overcome this issue, we propose a modification to the initialization of the different parameters. In our framework, we recap that the decoded weights used in a forward pass are obtained using
\begin{align}
    \bw=\text{reshape}(\bwt\bpsi)
    \label{eq:w}
\end{align}

where $\bwt$ is a matrix in $\mathbb{Z}^{\cin\cout \times l}$ and $\bpsi$ is a matrix in $\mathbb{Z}^{l \times l}$ (where $l=1$ for dense weights (and biases) while $l=K^2$ for convolutional weights).

Our goal is to initialize $\widehat{\bw}$ and $\bpsi$ such that the decoded weights $\bw$ follow He initialization \cite{he2015delving}. First, since $\widehat{\bw}$ is rounded to nearest integer (to obtain latent space weights $\bwt$), we assume its elements to be drawn from a uniform distribution in $[-b,b]$ where $b>0.5$ in order to enforce atleast some non-zero weights after rounding to nearest integer. Next, we take the elements of $\bpsi$ to be a normal distribution with mean $0$ and variance $v$. 

Assuming the parameters to be \iid, and $\text{Var}(X)$ denoting the variance of any individual element in matrix $X$,
\begin{align}
    \text{Var}(\bw) = l \times \text{Var}(\bpsi) \times \text{Var}(\bwt)
    \label{eq:wvar}
\end{align}

Assuming a RELU activation, with $f$ denoting the total number of channels (fan-in or fan-out) for a layer, LHS of \Cref{eq:wvar}, using the He initializer becomes $\frac{2}{f}$, RHS on the other hand can be obtained analytically
\begin{align}
    &\frac{2}{f} = l \times v \times \frac{(2b+1)^2-1}{12}  \nonumber\\
    \implies &b = \frac{\sqrt{\frac{24}{lvf}+1}-1}{2}, v = \frac{24}{lf\left((2b+1)^2-1\right)}
    \label{eq:bv}
\end{align}

\Cref{eq:bv} gives us a relationship between $b$ (defining the uniform distribution of $\widehat{\bw}$) and $v$ (defining the normal distribution of $\bpsi$). Note that $l$ and $f$ values are constant and known for each layer.

For a weight decoder corresponding to a parameter group, the maximum value of $f$ in that group enforces the smallest value of $b$ which should be above a minimum limit $b_{\min}$. Denoting $f_{\max}$ as the maximum fan-in or fan-out value for a parameter group, we get
\begin{align}
    &v = \frac{24}{lf_{\max}\left((2b_{\min}+1)^2-1\right)} \nonumber\\ 
    \implies
    &b = \frac{\sqrt{\frac{f_{\max}}{f}\left((2b_{\min}+1)^2-1\right)+1}-1}{2}
\end{align}
The hyperparameter $b_{\min}$ then refers to the minimum boundary any latent space parameter can take in the network. By calculating the values of $v$ based on $f_{\max},b_{\min}$ and $b$ for various parameters based on the corresponding value of $f$, we then initialize the elements of $\widehat{\bw}$ to be drawn from a uniform distribution in the interval $[-b,b]$ and elements of $\bpsi$ to be drawn from $\mathcal{N}(0,v)$.

Note that $f=f_{\max} \implies b = b_{\min}$ which shows that the minimum boundary corresponds to the layer with maximum channels (fan-in or fan-out) $f$.

By choosing an appropriate value of $b_{\min}$ we obtain good initial values of the gradient which allows the network to converge well as training progresses. $b_{\min}$ offers an intuitive way of initializing the discrete weights. Too small a value leads to most of the weights being set to zero while too large a value can lead to exploding gradients. In practice, we find that this initialization approach works well for Cifar experiments. For ImageNet experiments, we vary the variance of the weights manually and fix based on the best value.

\begin{figure}[t]
    \centering
    \includegraphics[width=0.6\linewidth]{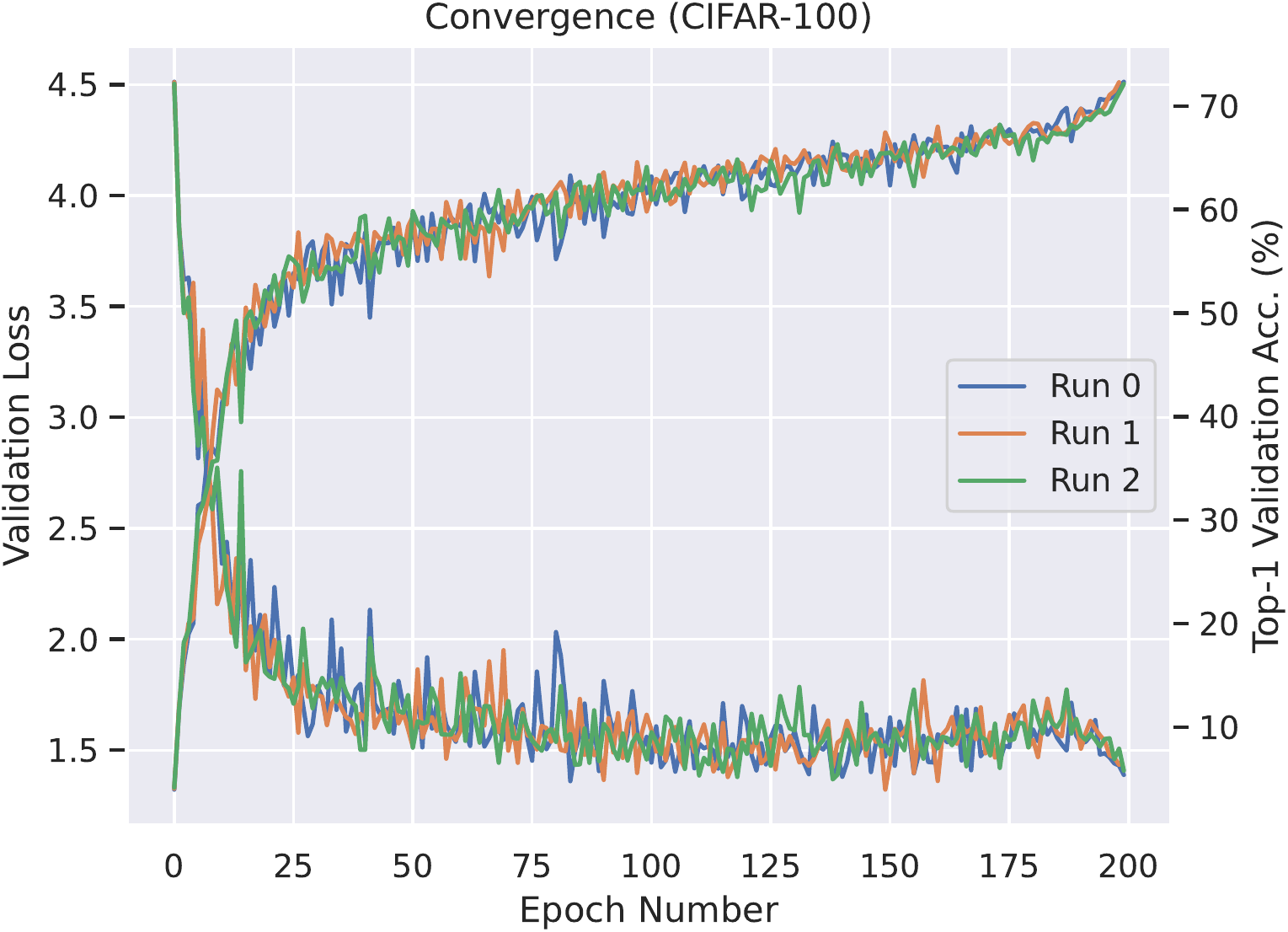}
    \caption{Convergence plots for 3 ResNet-20-4 runs on CIFAR-100. We see that loss (left axis) as well as top-1 validation accuracy (right axis) do not stabilize towards the end of training and respectively decrease/increase sharply suggesting that the training has not fully converged.}
    \label{supp_fig:convergence}
\end{figure}
\section{CIFAR-100 Convergence}
\label{supp_sec:convergence}
We analyze the convergence of 3 different runs for ResNet-20-4 trained on the CIFAR-100 dataset with varying values of $\lambdas$ and $\lambdau$. Results are shown in \cref{supp_fig:convergence} when trained for 200 epochs. We see that validation accuracy (on the right y-axis) continues to increase towards the end of training between 190-200 epochs. At the same time, validation loss (on the left y-axis) also decreases. This suggests that the model hasn't fully converged by the end of 200 epochs. We hypothesize that this is an artifact of the dataset as well as the cosine decay schedule where learning rate decreases drastically towards the end of training and is not maintained for longer for better convergence.

\begin{table}[!ht]
    \centering
    \caption{\textbf{Licenses of datasets}. }
    \label{tab:ds-license}
    \vspace{0.5em}
    \begin{tabular}{@{}l|cc@{}}
        \toprule
        Dataset & License \\
        \midrule
        CIFAR-10~\cite{krizhevsky2009learning} & MIT  \\
        CIFAR-100~\cite{krizhevsky2009learning}  & MIT \\
        ImageNet~\cite{deng2009imagenet} & BSD 3-Clause  \\
        \bottomrule
    \end{tabular}
\end{table}

\section{Code and License}
\label{supp_sec:license}

Table~\ref{tab:ds-license} lists all datasets we used and their licenses. 

Our code is available at \href{https://github.com/Sharath-girish/LilNetX}{https://github.com/Sharath-girish/LilNetX} with the MIT License.

\end{document}